%% file: iclr2025_conference.tex
\documentclass{article} 
\usepackage{iclr2025_conference,times}
\definecolor{citegreen}{RGB}{2, 142, 2}
\usepackage[pagebackref=false,breaklinks=true,colorlinks=true,citecolor=citegreen,bookmarks=false]{hyperref}
\input{math_commands.tex}

\usepackage{hyperref}
\usepackage{url}
\usepackage{booktabs}  
\usepackage{siunitx}   
\usepackage{tabularx}  
\usepackage{subcaption}    
\usepackage{cleveref}
\usepackage{fancyhdr}
\usepackage{graphicx}
\usepackage{booktabs}
\usepackage{multirow}

\usepackage{tikz}
\usepackage{enumitem}
\usepackage{tcolorbox}
\usepackage{colortbl} 
\usepackage{graphicx} 

\definecolor{myTableGreen}{rgb}{0.941,0.984,0.937}
\definecolor{myTableGrey}{gray}{0.937} 
\definecolor{myTableRed}{rgb}{0.996,0.945,0.945}

\title{xbench: Tracking Agents Productivity \\Scaling with Profession-Aligned Real-World Evaluations}

\author{Core Contributors \\
Kaiyuan Chen\thanks{Corresponding: chenky2022@gmail.com}, Yixin Ren, Yang Liu, Xiaobo Hu, Haotong Tian, Tianbao Xie, 
Fangfu Liu, \\ Haoye Zhang, 
Hongzhang Liu, Yuan Gong\thanks{Corresponding: ygong@hongshan.com} \And 
Contributors - listed alphabetically\\
Chen Sun, Han Hou,  Hui Yang, James Pan, Jianan Lou, Jiayi Mao, Jizheng Liu, Jinpeng Li, Kangyi Liu, \\Kenkun Liu, Rui Wang, Run Li, Tong Niu, Wenlong Zhang, Wenqi Yan, Xuanzheng Wang, \\Yuchen Zhang, Yi-Hsin Hung, Yuan Jiang, Zexuan Liu, Zihan Yin, Zijian Ma, Zhiwen Mo 
\And 
Affiliations - listed alphabetically\\
Carnegie Mellon University, Fudan University, Imperial College London, Massachusetts Institute of Technology, \\National University of Singapore, Peking University, Shanghai Jiao Tong University, Stanford University, \\The Chinese University of Hong Kong (Shenzhen), The Ohio State University, Tsinghua University, \\University of Chinese Academy of Sciences, University of Oxford, University of Pennsylvania, \\University of Science and Technology of China, University of Sydney, University of Toronto, Yale University
}

\iclrfinalcopy 

\begin{document}

\maketitle

\begin{abstract}

We introduce \textbf{xbench}, a dynamic, profession-aligned evaluation suite designed to bridge the gap between AI agent capabilities and real-world productivity. While existing benchmarks often focus on isolated technical skills, they may not accurately reflect the economic value agents deliver in professional settings. To address this, xbench targets commercially significant domains with evaluation tasks defined by industry professionals. Our framework creates metrics that strongly correlate with productivity value, enables prediction of Technology-Market Fit (TMF), and facilitates tracking of product capabilities over time.
As our initial implementations, we present two benchmarks: \textbf{Recruitment} and \textbf{Marketing}. For Recruitment, we collect 50 tasks from real-world headhunting business scenarios to evaluate agents' abilities in company mapping, information retrieval, and talent sourcing. For Marketing, we assess agents' ability to match influencers with advertiser needs, evaluating their performance across 50 advertiser requirements using a curated pool of 836 candidate influencers. We present initial evaluation results for leading contemporary agents, establishing a baseline for these professional domains. Our continuously updated evalsets and evaluations are available at \url{https://xbench.org/}.
\end{abstract}


\section{Introduction}

Breakthroughs in AI capabilities in long-text processing, multimodality, tool usage, and reasoning have catalyzed the rapid advancement of AI agents~\citep{gpt4o, claude3.7, gemini2.5, deepseekv3, r1, llama4, qwen2.5, qwen2.5vl, o3}. While chatbots excel at answering questions and providing information, agents can perform end-to-end task execution and workflow automation~\citep{liu2025advanceschallengesfoundationagents}, creating tangible productivity and commercial value~\citep{noy2023experimental, ju2025collaborating, su2024language-agent}. There is growing consensus that effective AI agent evaluation must align closely with real-world tasks. A series of high-quality evaluation sets~\citep{mmlu, mmlupro, gpqa, supergpqa, evalplus, math, pan2024webcanvas} have emerged in fields such as tool use, computer use, coding, and customer service, driving the rapid development of agents in these respective domains. However, current AI benchmarks do not predict the real-world economic impact, focusing on technical capabilities over business value. This limitation becomes critical as AI enters its evaluation-centric phase~\citep{ysymyth2025secondhalf}, necessitating domain-specific benchmarks that directly measure agent productivity and commercial utility in professional settings.

\begin{figure}[htbp]
    \centering
    \includegraphics[width=\linewidth]{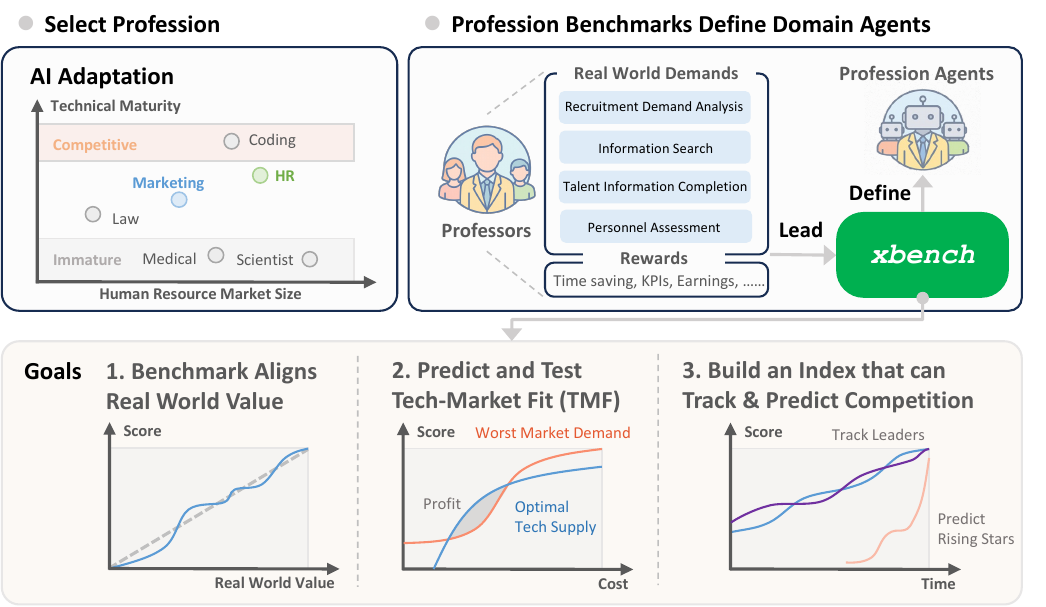}
    \caption{Profession-aligned evaluation define domain agents, predict Tech-Market Fit (TMF) and track competition of agent products.}
    \label{fig:proffesion-aligned}
\end{figure}

In this work, we introduce \textbf{xbench}, a profession-aligned series of evaluation sets for domain-specific agents. Specifically, we enable domain-specific agents to complete a series of real-world demands faced by domain experts, and use domain-oriented evaluation metrics to assess the performance of these agents. The core philosophy behind xbench's construction of profession-aligned evaluations is: (1) \textbf{Selecting evaluation domains based on market size and technological maturity.} The productivity value provided by an agent is a product of its technological maturity, the value of a single task, and the total demand for such tasks. Benchmarks can only provide an analysis of technological maturity, but market size guides us to find domain-specific agents that can scale commercially. Furthermore, domain-specific agents with intermediate technological maturity offer greater development value. High technological maturity often leads to intense competition, making it difficult for new, similar products to enter the market. Conversely, in overly immature domains, existing products may lack support or score too low, facing a period where the product cannot be brought to market; (2) \textbf{Domain experts lead domain-specific evaluations, and domain-specific evaluations define professional agents.}
The real-world demands, work environments, and feedback from industry experts constitute the most ideal evaluation tasks. The evaluation set should also align as closely as possible with the distribution of expert demands and cover the entire workflow for task completion.
A profession-aligned evaluation system can further define the ideal state of a domain-specific agent product, serving as a direction for agent iteration while also constraining the upper limits of agent capabilities.

Profession-aligned evaluation metrics have the following objectives:
\begin{itemize}[leftmargin=*]
    \item \textbf{Strong correlation between agent scores and real-world value:} The market should be able to price agent product services based on these metrics. Improvements in benchmark scores should translate to enhancements in the domain-specific experience.
    \item \textbf{Predicting and validating Technology-Market Fit (TMF):} This involves analyzing benchmark metrics against costs. For the market pricing of each evaluation set metric, we can statistically determine the worst-case market acceptance curve, as shown in \cref{fig:proffesion-aligned}. Through evaluation, we can determine the optimal technology feasibility curve. The intersection of these two curves signifies TMF achievement, and the area of intersection represents the value generated by the agent per unit task.
    \item \textbf{Building xbench-Index to track and predict competition among agent products through long-term updated evaluations:} We can track alternate leading products, and we also hope to discover rising stars whose capabilities rapidly improve in the short term.
\end{itemize}

The first batch of xbench covers two professional domains with high market value and moderate technological fit: Recruitment and Marketing. The recruitment agent evaluation focuses on scenarios where headhunters recommend candidates. The agent needs to complete 50 tasks from business scenarios of headhunters, including company mapping,  information retrieval and talent sourcing, thereby assisting professional headhunters from demand to candidate recommendation. The marketing agent evaluation requires finding suitable influencers for advertising on video and social media platforms based on product information and promotion needs provided by clients. We constructed the evaluation set using 50 real client demands and profile data from 836 selected influencers for promotion, covering marketing tasks for advertising in e-commerce, apps, and gaming domains.


We believe that xbench will contribute to discovering, defining, and predicting outstanding domain-specific agent products, and help analyze the competitive advantages and technological barriers among agent teams in the same domain. Our contributions include:
\begin{itemize}[leftmargin=*]
    \item \textbf{xbench:} A profession-aligned evaluation series, with the first batch providing evaluation methods for domain-specific agents in Recruitment and Marketing.
    \item \textbf{Evaluations of agents on xbench:} We will continuously report our evaluation results of the professional capabilities of specialized agents and general-purpose agents.
    \item \textbf{Long-term updated evaluations:} We will continuously update our evaluations and assess agent products, capturing the dynamic growth in agent application capabilities.
\end{itemize}

\section{AI Capability Centric v.s. Profession-Aligned}

\begin{figure}
    \centering
    \includegraphics[width=\linewidth]{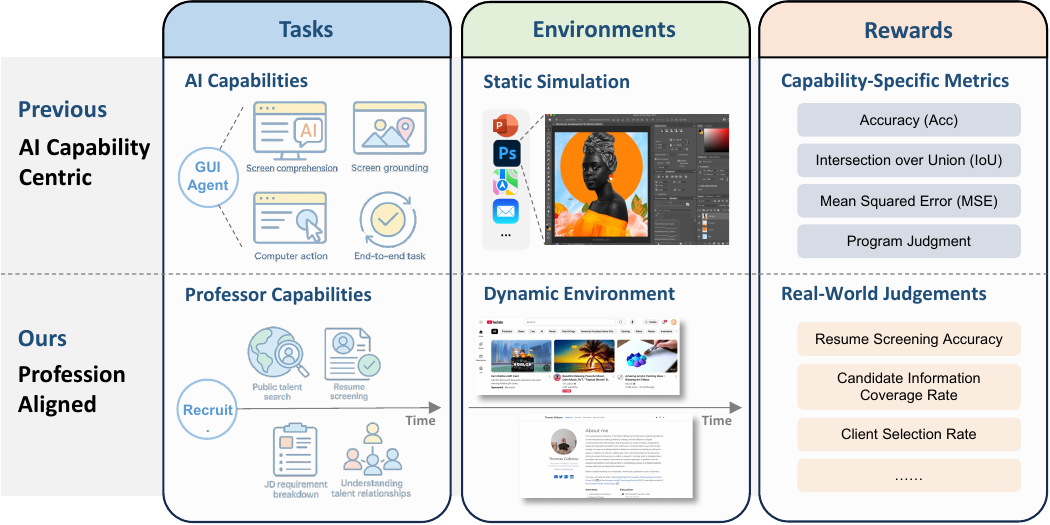}
    \caption{Difference between AI-capability-centric and profession-aligned benchmarks}
    \label{fig:framework}
\end{figure}

Artificial intelligence is experiencing exponential growth in the length of performable tasks. According to research by METR~\citep{measuring-ai-ability-to-complete-long-tasks}, this capability approximately doubles every seven months. However, most existing evaluations still focus on short, simulated tasks. More specifically, they are typically constructed based on specific agent capabilities that prioritize a dimension to be evaluated, such as coding, database operations, or GUI computer use. The main limitation of these evaluations is that the indicators tend to saturate, making it difficult to distinguish the deep capabilities of different models. Therefore, there is a gap between AI capability-centered evaluation sets and the performance of real domain-specific agents. 

To fill the gap, it is essential to build a series of new evaluations, aiming to (1) expand the assessments of the upper limits of AGI capabilities; and (2) more accurately measure the productivity value provided by AI in real-world applications, including economic benefits, innovation, and complex problems.  Such evaluation systems are crucial for guiding AI to develop in a more practically valuable direction. Towards this end, we propose a profession-aligned evaluation construction method for professional agents and build a dynamically maintained series of evaluation sets, hoping to track, guide, and predict the performance of professional agent products in the long term. As shown in \cref{fig:framework}, our proposed profession-aligned evaluations differ from AI capability-centered evaluations in terms of evaluation direction, task distribution, environment, and feedback mechanisms:





\begin{itemize}[leftmargin=*]
    \item \textbf{Evaluation Direction.} AI capability-centric evaluation focuses on creating distinct scenarios to test tasks where existing models and systems exhibit deficiencies or lack mastery. Profession-aligned evaluation, on the other hand, prioritizes scenarios that offer significant productivity improvements and commercial value as its primary evaluation focus.

    \item \textbf{Task Distribution.} Capability-centric evaluation aims to encompass diverse tasks across multiple domains and task types to maximize task variety. In contrast, profession-aligned evaluation focuses on real-world demands from domain experts, ensuring each task delivers productivity or business value and aligns metrics with practical outcomes. To achieve this, we collaborate with domain experts to co-design the evaluation system and tasks.
    

    \item \textbf{Environment.} To ensure reproducible and comparable task test results, existing evaluations often use static or simulated environments for task construction, which inevitably creates a Sim-to-Real Gap. In contrast, the profession-aligned evaluation seeks maximum alignment between the evaluation and real-world work environments, requiring interaction with dynamic tools, websites, or systems. To maintain comparability of evaluation metrics across different periods, we propose developing a temporally consistent index for agent products, as discussed in Section~\ref{sec:xbench_index}.

    \item \textbf{Feedback Mechanism.} Capability-centric evaluation metrics focus on task completion and pursue metric correctness, which may exclude tasks that can affect the metric accuracy. In contrast, profession-aligned evaluation aligns evaluation scores with key business indicators. For tasks challenging to evaluate objectively, we employ LLM-as-a-judge based on professional rubrics for scoring. 
\end{itemize}

\section{Building xbench}

We select recruitment and marketing as the first two domains for the profession-aligned xbench evaluation sets. Both domains involve a significant amount of human effort in information collection and analysis, making them suitable for DeepResearch-type AI agents. The results delivered in these areas can directly generate business value. We have co-constructed the recruitment and marketing evaluation tasks with professional headhunters and marketing enterprises, respectively, which have accumulated substantial historical business data.

\subsection{Live Professional Demands Define Evaluation Tasks}

The construction of xbench is guided by three core principles:

\begin{itemize}[leftmargin=*]
    \item \textbf{Evaluation is defined by demand.} When building an evaluation set for a profession, the priority is to establish its business processes and demand classifications. We conduct extensive interviews with field experts to map how they allocate their weekly work time across different tasks. Experts also assess the importance of these tasks to create a detailed breakdown of domain work. For each task, we analyze its testability and feasibility from a current technological perspective, categorizing them as outlined in \cref{tab: tab1}. We prioritize focusing on the evaluable components of tasks. Some inevaluable tasks can be transformed into evaluable ones through simulation.

    \item \textbf{Evaluation tasks are collected ``live" from expert business operations as they arise over time.} Real business demands are not "created" by devising questions but are accumulated and collected during business progress. xbench tracks the real business of corresponding enterprises in each domain, organizing the latest business into evaluable tasks for dynamic updates. At the same time, expert business operations often have life cycles. For example, in marketing, the popularity of influencers varies over time; they might suddenly become viral, stop updating, or gradually fade. In such cases, the evaluation environment for a marketing task changes. We label each task as either static or dynamic. For dynamic evaluations, we continuously collect tasks from real business operations that are most relevant to the current market environment.

    \item \textbf{Domain value drives evaluation objectives.} Each business task has a predefined Technology-Market Fit (TMF) target. Tasks' difficulty aims to align with reality rather than becoming progressively harder. For each task, we will note the time it takes for a human expert to complete it and estimate the value of each task based on the corresponding expert salary.
\end{itemize}

\begin{table}[!t]
\caption{Task Categorization by Feasibility and Evaluability}
\centering
\small
\begin{tabularx}{\linewidth}{@{} l X X @{}}
\toprule
 & \textbf{Evaluable} & \textbf{Inevaluable} \\
\midrule
\textbf{Feasible}
 & AI can complete the task, producing a definitive and scorable result. 
 & AI can perform the task, but the output is subjective and lacks a clear scoring standard.  \\
\addlinespace 
\textbf{Infeasible}
 & Tasks currently beyond the AI's technical capabilities (e.g., I/O), but whose hypothetical outcome could still be clearly evaluated. .
 & AI can neither perform the task nor could its outcome be effectively evaluated, often due to complex real-world interactions with long validation cycles. \\
\bottomrule
\end{tabularx}
\label{tab: tab1}
\end{table}



\subsection{Benchmarking Recruitment} 

 The global recruitment market is valued at approximately \$700 billion. Within this, external recruitment, as part of Recruitment Process Outsourcing (RPO), is rapidly growing. In 2024, the external recruitment market size is about \$11.9 billion and is projected to reach \$37 billion by 2031, with a compound annual growth rate (CAGR) of 17.6\%. Against the backdrop of increasingly complex enterprise recruitment needs, external recruitment enhances hiring efficiency and effectiveness through professional services. 

 The core tasks of external recruitment are divided into two main modules: information collection and communication interaction. 

 \begin{itemize}[leftmargin=*] 
 	\item \textbf{Information Collection:} This includes candidate information search, resume screening, and talent persona construction, typically occupying over 50\% of a recruiter's time and effort. The recruitment benchmark evaluates an agent's ability to use public information to progress from client demand to candidate recommendation. Each effectively recommended candidate directly benefits the client. 
 	\item \textbf{Communication Interaction:} Building on information collection, communication interaction is a critical skill for senior recruiters. It includes confirming requirements with clients, discussing career development with candidates, salary negotiations, and obtaining "below-the-surface" information. Evaluating these skills involves writing operations and feedback with the external world, such as real-time communication, for which an automated testing foundation is currently lacking. Therefore, we temporarily focus on quantifiable information collection tasks. 
 \end{itemize} 

 \begin{table}[htbp] 
 \caption{Structured breakdown of recruitment work tasks} 
 \centering 
 \small 
 \begin{tabular}{@{} l l c c @{}} 
 \toprule 
 \textbf{Evaluation Category} & \textbf{Work Type} & \textbf{Feasible} & \textbf{Evaluable} \\ 
 \midrule 
 \multirow{2}{*}{Recruitment Demand Analysis} & JD Requirement Decomposition & Yes & Yes \\ 
 	& Talent Persona Positioning & Yes & Yes \\ 
 \midrule 
 \multirow{3}{*}{Information Collection} & Public Talent Search & Yes & Yes \\ 
 	& Personnel Relationship Graph & Yes & Yes \\ 
 	& Company Org Structure Analysis & No & Yes \\ 
 \midrule 
 \multirow{2}{*}{Talent Information Completion} & Candidate Experience Completion & Yes & Yes \\ 
 	& Cold Call Information Gathering & Yes & No \\ 
 \midrule 
 \multirow{3}{*}{Personnel Assessment} & Resume Screening & Yes & Yes \\ 
 	& Interview & No & No \\ 
 	& Career Development Guidance & Yes & No \\ 
 \bottomrule 
 \end{tabular} 
 \end{table} 

 Considering feasibility and testability, the current recruitment benchmark includes tasks such as talent persona positioning (mapping corresponding companies according to recruitment demands), candidate experience completion, and public talent search. These tasks assess multiple agent capabilities, including industry knowledge and talent search. Additionally, some tasks require the agent to provide precise talent information or candidate screening, which also evaluates the agent's trustworthiness. 

 \paragraph{Task Design and Examples} 
 We have designed evaluation task collection and assessment schemes for three main task types, collaborating with professional headhunters to curate evaluation tasks from their business operations and experience. Here, we present the task design methods and examples for these three categories. 

 \begin{itemize}[leftmargin=*] 
    \item \textbf{Company Mapping:} The agent needs to identify suitable schools, companies, teams, etc., to source talent based on a Job Description (JD). A skillful headhunter can quickly and accurately pinpoint the required teams for a recruitment demand, using this prior knowledge to enhance efficiency. The task input is a job description, typically containing information about the company's industry and business, as well as background, competency, and trait requirements for the candidate, as shown in \cref{tab:case_company_mapping}. The headhunter also provides the ideal teams or companies for the position, ensuring the provided verification answers are consensual and comprehensive based on industry experience. 
    \item \textbf{People-to-Info:} The agent receives the name of a target individual along with partial information and is required to complete the person's professional history. The annotator provides a candidate and their public information, which is divided into two parts: one part is included in the prompt to specify the individual, and the other is used to create verification questions. The agent should produce a comprehensive talent profile, and a judge model will score the collected results by checking if they correctly answer the verification questions. 
    \item \textbf{Info-to-People:} This task requires the agent to find a specific public figure based on a set of constraints. The annotator starts with a target individual or a group of individuals with common characteristics and progressively builds constraints to ensure the correct answer aligns with the target. The annotator must ensure that the answers are accurate and comprehensive. 
 \end{itemize} 

\begin{table}[ht]
\caption{An example of Company Mapping task. For the complete example, please refer to \cref{tab:case_full_company_mapping}}
\begin{small}
\resizebox{\linewidth}{!}{
    \begin{tabular}{@{}ll@{}}
    \toprule
    \textbf{Task Type} &
      Company Mapping \\ \midrule
    \textbf{\begin{tabular}[c]{@{}l@{}}Task\\ Description\end{tabular}} &
      \begin{tabular}[c]{@{}l@{}}Target Candidate Profile Analysis Based on Job Requirements\\    \textbf{Job Analysis:}\\ This role is an Influencer Marketing Specialist for an ACG (Animation, Comics, \\ Games) AIGC (AI Generated Content) product. The core requirement is ...\\ \textbf{Target Talent Profile:}\\ (1) Core Background Requirements:\\ - Bachelor's degree or higher, with a minimum of three years of experience in influ-\\encer marketing...\\ (2) Professional Competencies:\\- Proven track record of successful viral influencer marketing campaigns...      \\ (3) Personal Traits:\\- Extremely self-driven with an entrepreneurial mindset...\end{tabular} \\ \midrule
    \textbf{\begin{tabular}[c]{@{}l@{}}Verifier\\ Answers\end{tabular}} &
      \begin{tabular}[c]{@{}l@{}}Seasoned professionals J and K from marketing agencies, specialized MCN organiz-\\ations G, H, and I, in addition to the marketing and growth divisions E and F of social\\ media platforms, e-commerce platforms C and D, and short-video platforms A and B.\end{tabular} \\ \bottomrule
    \end{tabular}
}
\end{small}
\label{tab:case_company_mapping}
\end{table}

\begin{table}[ht]
\caption{A brief example of People-to-Info task. For the complete example, please refer to \cref{tab:case_full_people_to_info}}
\resizebox{\linewidth}{!}{
    \begin{tabular}{@{}ll@{}}
    \toprule
    \textbf{Task Type} &
      People-to-Info \\ \midrule
    \textbf{\begin{tabular}[c]{@{}l@{}}Task\\ Discription\end{tabular}} &
      \begin{tabular}[c]{@{}l@{}}Seeking comprehensive information for Ming Li, Backend Developer at Company A.     \\ \textbf{Reference Information:}     \\ (1) Work Experience     \\     - August 2023 – Present: Company A, Java Backend Developer ...\\ (2) Education Background\\     - 2014 – 2018: University A, Bachelor of Science in Computer Science\end{tabular} \\ \midrule
    \textbf{\begin{tabular}[c]{@{}l@{}}Verifier\\ Questions\end{tabular}} &
      \begin{tabular}[c]{@{}l@{}}\textbf{Question 1:} What were his key technical achievements during his tenure at Company A?\\ ......\end{tabular} \\ \midrule
    \textbf{\begin{tabular}[c]{@{}l@{}}Verifier\\ Answers\end{tabular}} &
      \begin{tabular}[c]{@{}l@{}}\textbf{Answer 1:}\\      1) Responsible for the backend optimization of the short-form video recommendation\\      system, which increased the click-through rate by 18\%.\\      2) Developed real-time interactive features for live streaming rooms, supporting tens \\      of millions of concurrent users.\\      3) Constructed a user profiling and analytics platform.\\ ......\end{tabular} \\ 
    \bottomrule
    \end{tabular}
}
\label{tab:case_people_to_info}
\end{table}

\begin{table}[ht]
\caption{An example of Info-to-People task.}
\resizebox{\linewidth}{!}{
    \begin{tabular}{@{}ll@{}}
    \toprule
    \textbf{Task Type} &
      Info-to-People \\ \midrule
    \textbf{\begin{tabular}[c]{@{}l@{}}Task\\ Description\end{tabular}} &
      \begin{tabular}[c]{@{}l@{}}Seeking General Manager, Industry A\\ \\ The ideal candidate will have previously held the top leadership position (e.g., CEO, \\President, or General Manager) within a relevant enterprise in Industry A, operating \\within mainland China.\\ \\ A critical requirement is that the candidate's former or current company possesses \\domestic research and development (R\&D) capabilities within China. The company \\should not be solely a distribution entity for overseas products.\end{tabular} \\ \midrule
    \textbf{\begin{tabular}[c]{@{}l@{}}Verifier\\ Answers\end{tabular}} &
      Person A, Person B \\ \bottomrule
    \end{tabular}
}
\label{tab:case_info_to_people}
\end{table}

 \paragraph{Evaluation Process} 

 The recruitment benchmark uses open-ended responses combined with an LLM Judge, which analyzes the responses and assigns a score from 1 to 5, with higher scores being better. A score of 5 indicates that the task is completed accurately and without hallucinations, while a score of 1 signifies a completely incorrect answer with hallucinations. The LLM Judge employs a Chain-of-Thought format, first providing its analysis and then the final score. In our final report, we linearly map the 5-point scale to a 0-100 point scale for better analysis and comparison. 

\begin{figure}[htbp]
    \centering 
    \includegraphics[width=0.9\linewidth]{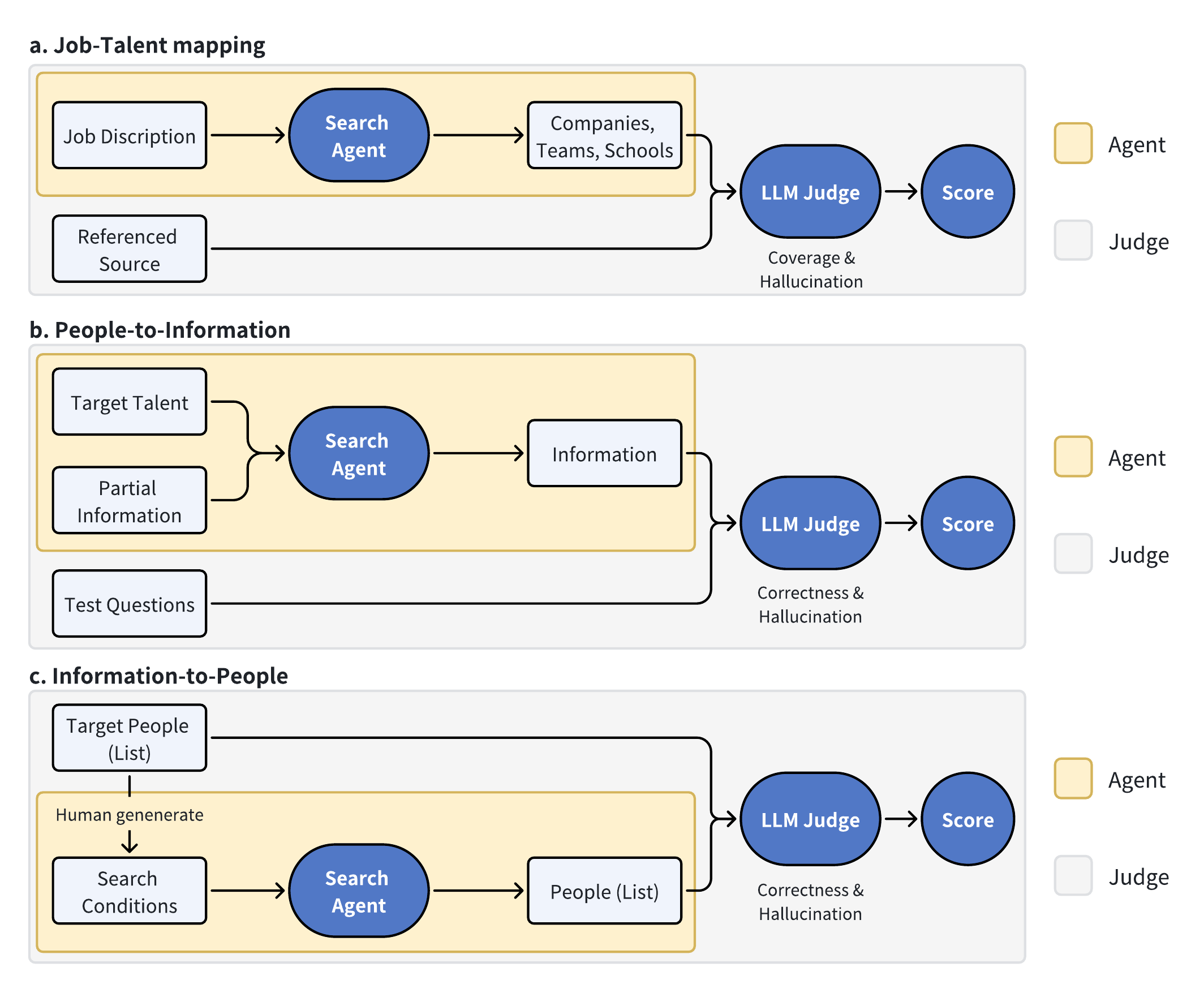} 
    \caption{Evaluation pipelines for recruitment tasks.} 
    \label{fig:recruitment_pipeline} 
 \end{figure} 
\Cref{fig:recruitment_pipeline} illustrates the specific evaluation process. The light-yellow areas represent the model's inference results, while the gray areas depict the judgment process. The prompts used for evaluation and judgment are provided in the appendix~\ref{appdix:recruit_prompts}. \Cref{fig:recruitment_pipeline}a describes the Company Mapping task, where the search agent provides search results for companies, teams, or schools. The LLM Judge then matches these against the annotated sources and performs hallucination detection to give a final score. \Cref{fig:recruitment_pipeline}b corresponds to the People-to-Info task, where the search agent gathers as much public experience information as possible based on partial information about a candidate. The Judge model uses pre-prepared verification questions to check if the collected information is sufficient and considers the correctness of the answers, whether the model has hallucinated non-existent experiences, or collected information about other individuals, to provide a final score. \Cref{fig:recruitment_pipeline}c describes the Info-to-People process, where the model needs to find as many target individuals as possible based on a series of search conditions, without being pre-informed of the number of people to find. The Judge model will then score the results based on correctness and whether the model has provided hallucinatory information. 

\begin{figure}
     \centering
     \includegraphics[height=4.5cm]{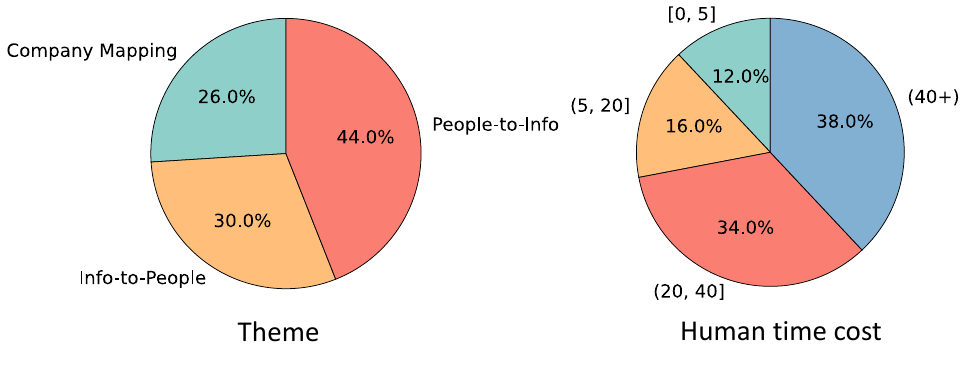}
     \caption{Task distribution across recruitment tasks' themes and human time cost (in minutes).}
     \label{fig:recruitment_dist}
 \end{figure}

 \paragraph{Task Distribution} 
 We collected 50 real business cases faced by headhunters and confirmed that all tasks can be solved using publicly available information, and the information collected about individuals is all self-disclosed from reliable media or websites. As shown in \cref{fig:recruitment_dist}, the three task types, Company Mapping, Info-to-People, and People-to-Info, account for 44\%, 30\%, and 26\% of the evaluation set, respectively. We also annotated the time required for each task in minutes: 12\% of tasks take 0 to 5 minutes, 16\% take 5 to 20 minutes, 34\% take 20 to 40 minutes, and 38\% take over 40 minutes. 



\subsection{Benchmarking Marketing}

The global online marketing market is approximately \$650 billion. Within this, marketing strategy formulation accounts for \$10 billion, influencer marketing for \$35 billion, Public Relations(PR) marketing for \$20 billion, and social media operations/brand management for \$10 billion. The largest segment is performance advertising at \$550 billion. This marketing benchmark primarily selects the influencer marketing segment, which faces issues such as being labor-intensive, having low matching efficiency, and non-transparent pricing.

In the influencer marketing field, many Multi-Channel Networks (MCNs) provide intermediary services. The workflow for their operational staff includes client demand communication, search, influencer matching, pricing, script polishing, content review, and content publishing. Most of this work is still done manually.

\begin{table}[htbp]
\caption{Structured breakdown of influencer marketing work tasks}
\centering
\small
\begin{tabular}{@{} l l c c @{}}
\toprule
\textbf{Evaluation Category} & \textbf{Work Type} & \textbf{Feasible} & \textbf{Evaluable} \\
\midrule
\multirow{2}{*}{Client Demand Communication} & Client Demand Analysis & Yes & Yes \\
 & Business Communication & No & No \\
\midrule
\multirow{3}{*}{Influencer (KOL) Search} & Influencer Search & Yes & Yes \\
 & Influencer Filtering & Yes & Yes \\
 & Influencer Traffic Prediction & Yes & Yes \\
\midrule
\multirow{2}{*}{Influencer (KOL) Match} & Ad Demand and Plan Negotiation & No & No \\
 & Placement Price Negotiation & No & No \\
\midrule
\multirow{3}{*}{Ad Customization} & Ad Copy Design & Yes & No \\
 & Video Production & Yes & No \\
 & Video Review & Yes & Yes \\
\midrule
\multirow{2}{*}{Content Publishing} & Propagation Effect Monitoring & Yes & No \\
 & Strategy Adjustment & Yes & Yes \\
\bottomrule
\end{tabular}
\end{table}

Due to data availability and evaluability, the current marketing benchmark focuses on a complex task category: Influencer Search. Other aspects, such as the communication involved in client demand and influencer matching, as well as propagation effect monitoring and strategy adjustment, remain under exploration for feasible evaluation schemes due to their dynamic and delayed nature.

In Influencer Search tasks, personnel screen suitable influencers based on client product information. The client then selects candidates from the list submitted. The marketing benchmark uses 50 real promotion demand cases from marketing companies' history, along with the ideal influencers ultimately selected by clients or experts. We evaluate the end-to-end match between the list of influencers recommended by the agent and the real list. An LLM Judge is used to determine if the influencers recommended by the marketing agent align with the ideal persona, providing an estimated re-selection rate for the agent's results. A higher estimated match implies a higher actual client re-selection rate.

\paragraph{Task Design and Examples}
We collaborated with senior operations staff from frontline marketing firms and, with company permission, collected anonymized real-world business requirements. \Cref{tab:campaign-requirements} presents an example of a marketing campaign brief, which includes basic product information, key selling points provided by the brand, and campaign-related details such as objectives, budget, and requirements for the desired influencers (e.g., quantity and persona).

For each promotional campaign, the historical data includes the list of influencers recommended by the marketing operations team to the client, as well as the final list of influencers the client actually selected. The overlap and differences between the initial recommendations and the client's final selections help us construct a persona of the ``ideal influencer," which serves as the ground truth for evaluation. Each campaign brief is typically associated with a list of 2 to 30 selected influencers used to build this ideal persona.

\begin{table}[htbp]
\centering
\caption{Influencer marketing campaign requirements.}
\label{tab:campaign-requirements}
\begin{tabular}{
    >{\raggedright\arraybackslash}p{3.5cm} 
    >{\raggedright\arraybackslash}p{10cm}  
}
\toprule
\textbf{Dimension} & \textbf{Content} \\
\midrule
Basic Information &
\textbf{Product Name:} X Brand Dual Basket Air Fryer \\
& \textbf{Client:} X Electronics Brand \\
& \textbf{Product Type:} 8-in-1 Smart Air Fryer \\
& \textbf{Launch Time:} October Release \\
\midrule
Product Highlights &
• Dual basket design, 10-quart large capacity \\
& • 8+ preset functions: air fry, grill, dehydrate, bake, roast, broil, reheat, toast \\
& • Smart sync finish technology \\
& • Wi-Fi connectivity, voice assistant support \\
& • 50+ app recipes \\
& • Transparent viewing window design \\
\midrule
Campaign Objectives &
Drive off-site traffic with links, increase product exposure, boost sales \\
\midrule
Budget \& Distribution &
\textbf{Total Budget:} \$50,000 USD \\
& \textbf{Target Region:} North America \\
& \textbf{Platform Allocation:} YouTube (40\%) / TikTok (30\%) / Instagram (30\%) \\
\midrule
Influencer Requirements &
\textbf{Quantity:} 15 influencers \\
& \textbf{Types:} Food/health, kitchen tech review, air fryer buying guide creators \\
& \textbf{Audience Profile:} Fitness enthusiasts, low-oil healthy diet followers, busy families,
child health conscious parents, party snack lovers \\
& \textbf{Requirements:} Optimal resource allocation within budget \\
\bottomrule
\end{tabular}
\end{table}

\paragraph{Evaluation Process}

\begin{figure}[htbp]
    \centering
    \includegraphics[width=\linewidth]{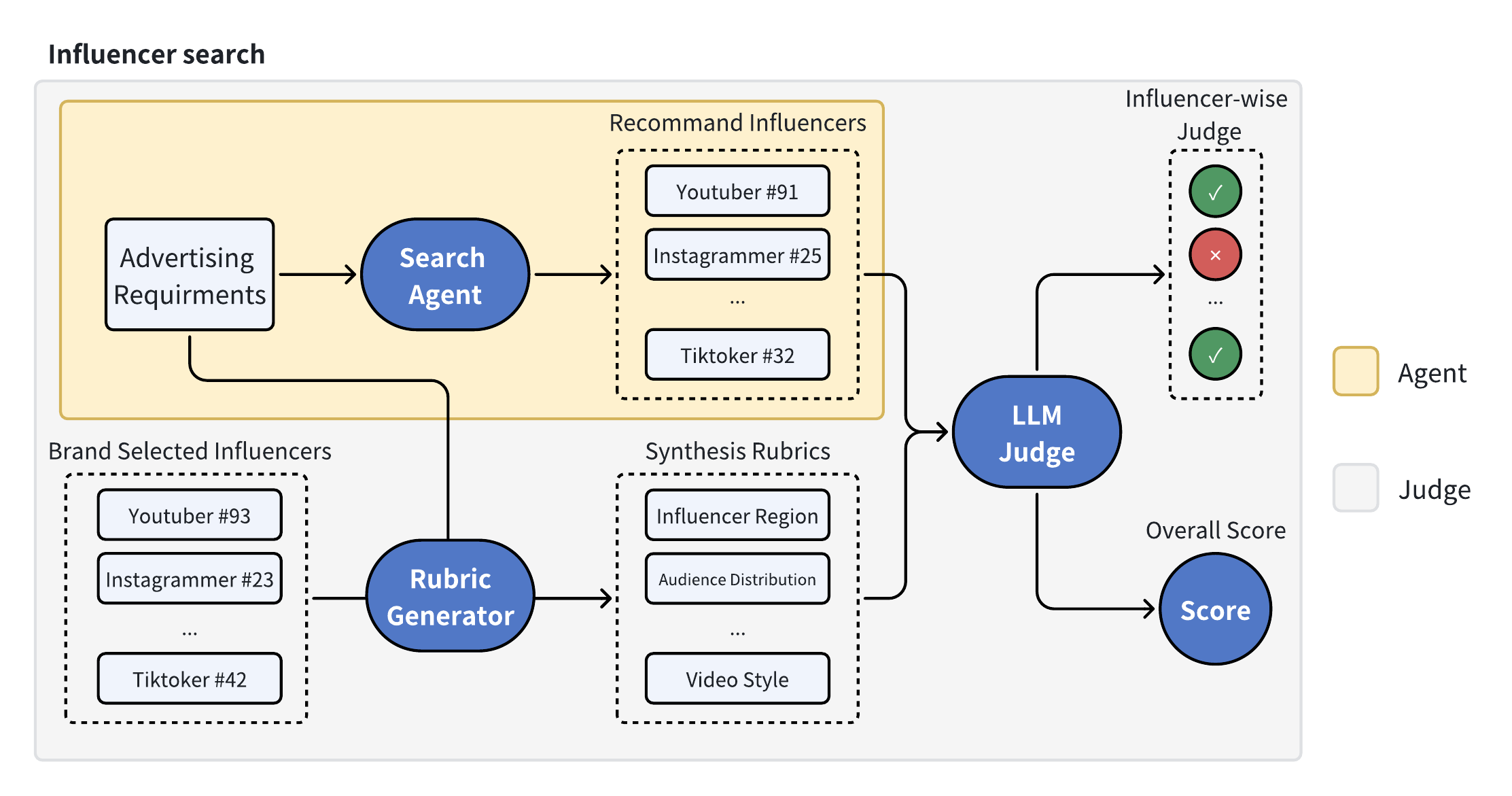}
    \caption{The evaluation pipeline for the Influencer Search task in marketing benchmark.}
    \label{fig:marketing_eval_pipeline}
\end{figure}

\Cref{fig:marketing_eval_pipeline} illustrates the evaluation scheme for the Influencer Search task, following the same diagrammatic style as the previous section. Search agents are required to find ideal influencers on platforms like YouTube, Instagram, and TikTok based on the campaign requirements. Agents can use any available tools, including but not limited to Search, Long Thinking, Deep Research, and Browser-use, to produce a final list of recommended influencers. The tasks and evaluation prompts for this process are detailed in the \cref{appdix:recruit_prompts}.

We have built an automated system to collect influencer data, including basic profiles, recent video performance, audience demographics, and past brand collaborations, summarizing this information in text format. We use this system to analyze the client-selected influencers to understand the "ideal persona" and to assess the suitability of the influencers recommended by the agent.

The evaluation process consists of two main stages: first, generating a detailed scoring rubric for each task, and second, using that rubric to score each influencer recommended by the agent, ultimately aggregating these into a final task score.
The \textbf{Rubric Generator} uses an LLM to summarize the profile of each influencer selected by the client. It then combines these summaries with the original campaign brief to generate a detailed, itemized rubric of the ideal influencer persona.

\begin{figure}
    \centering
    \includegraphics[height=4.5cm]{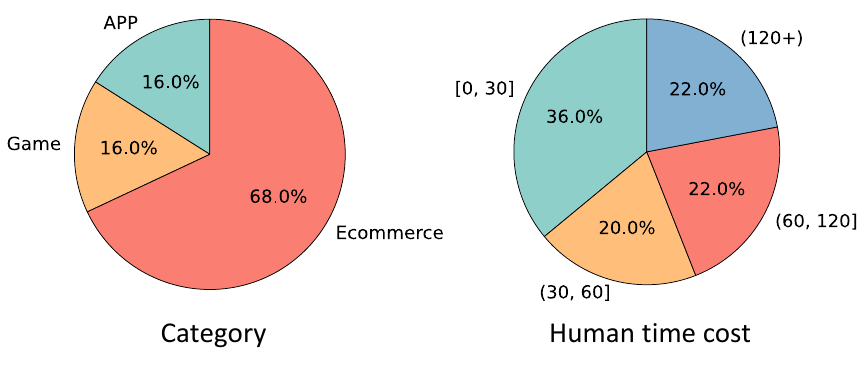}
    \caption{Task distribution across recruitment tasks' categories and human time cost (in minutes).}
    \label{fig:marketing_dist}
\end{figure}

\paragraph{Task Distribution}
We collected 50 advertising campaign tasks and anonymized them by replacing specific company and product names with general industry and category information. This approach maintains the integrity of the evaluation task while ensuring client data confidentiality. The evaluation tasks are categorized by the client's industry: App (68\%), Game (16\%), and E-commerce (16\%), shown in \cref{fig:marketing_dist}. Different industries typically require influencers from distinct fields and involve different promotion strategies. We also annotated the manual effort required for each task: 36\% take 0 to 30 minutes, 20\% take 30 to 60 minutes, 22\% take 60 to 120 minutes, and 22\% require over 120 minutes.

\section{Evalutions}

\subsection{Setups}
All evaluation tasks require web connection at a minimum, though the ability to use additional tools may provide incremental benefits. From an evaluation perspective, we do not impose restrictions on the solution architecture and only assess the final results. We evaluated a range of agents and models, including o3 (OpenAI)~\citep{o3}, o4-mini-high (OpenAI), GPT-4o (OpenAI)~\citep{gpt4o}, Claude-3.7-Sonnet~\citep{claude3.7}, Gemini-2.5-Pro, Gemini-2.5-Flash~\citep{gemini2.5}, Perplexity-Search, Perplexity-Research, and Deepseek R1~\citep{guo2025deepseek}. The marketing metrics for Deepseek R1 are not reported, as it lacks access to necessary search sources like YouTube.

All evaluations are conducted using the web-based interfaces of these products with their internet search capabilities enabled. The testing period is concentrated in May 2025, during which there were no significant version updates to the products being tested. The results of this evaluation may be influenced by future updates to the agent versions and changes in the internet environment, a point we address in the Discussion section~\ref{sec:xbench_index}. 

All LLM Judge scoring is performed using the Gemini-2.5-Flash model. 
We plan to subsequently update our analysis with more details on metric stability, alignment with human judgment, and consistency across different judge models.

\subsection{Benchmarks}
\begin{figure}[htbp]
    \centering
    \includegraphics[width=\linewidth]{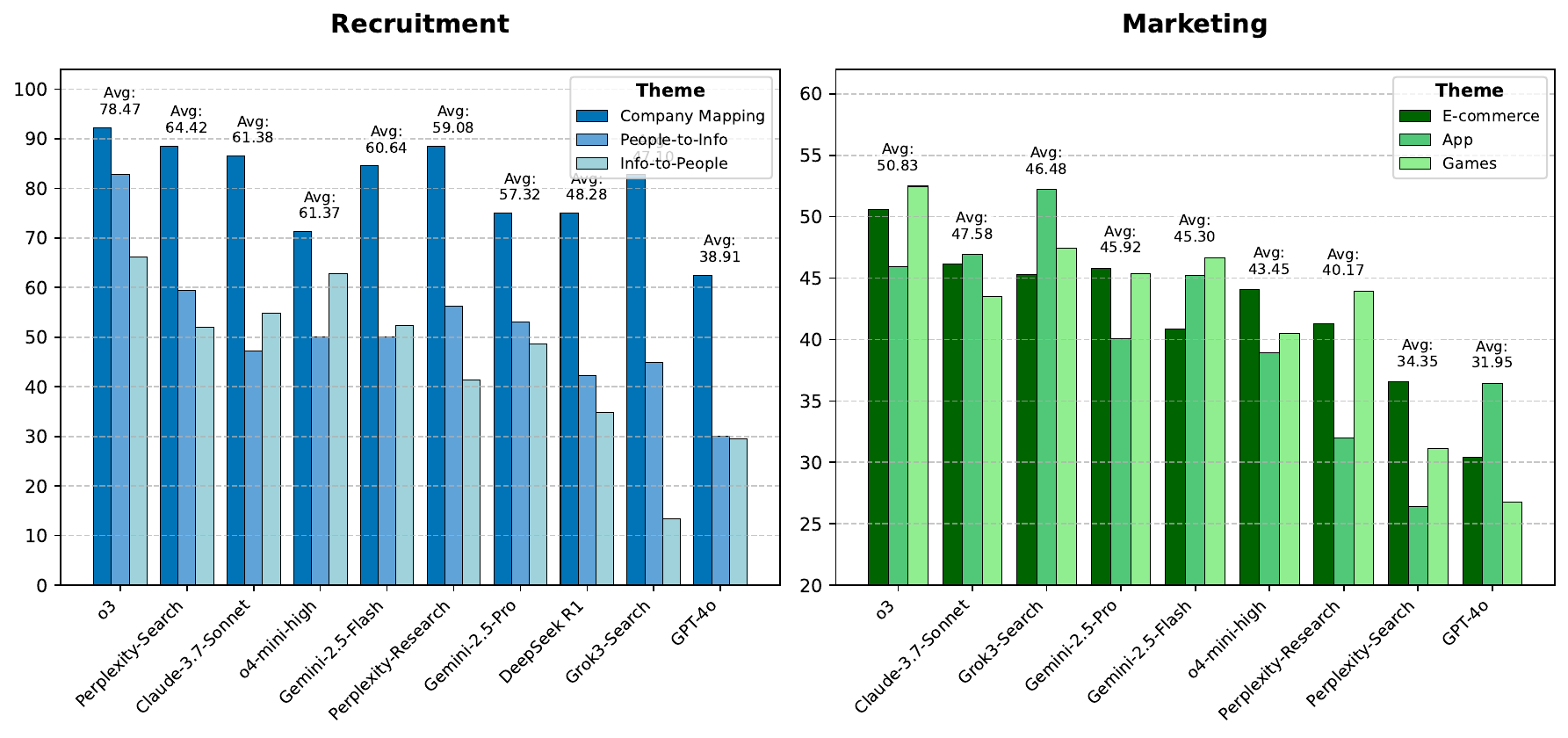}
    \caption{Score chart of xbench by themes.}
    \label{fig:benchmark_theme}
\end{figure}

We report the xbench evaluation results for Recruitment and Marketing in \cref{tab:recruitment_benchmark} and \cref{tab:marketing_benchmark}, respectively. The score leaderboards for different themes are also depicted in the \cref{fig:benchmark_theme}.

The results show that o3 ranks first on both benchmarks, suggesting that its end-to-end reinforcement learning, combined with strong search capabilities, significantly enhances its domain-specific performance. In contrast, GPT-4o, which tends to provide shorter responses, ranked last. Notably, a larger model size does not guarantee a significant advantage, as the performance of Gemini-2.5-Pro and Gemini-2.5-Flash was comparable. It should also be noted that since Gemini-2.5-Flash was used as the judge model, there is a potential for bias that may lead to an overestimation of its own performance.

Interestingly, Perplexity-Search even outperformed its Research version on the Recruitment tasks. This may imply that the extended research process can introduce a higher rate of hallucinations. Finally, despite Deepseek R1's excellent performance on math and code benchmarks, its lack of adaptation for search-centric tasks resulted in lower performance here.

\begin{table}[htbp]
\centering
\caption{Evaluation results on marketing}
\label{tab:benchmark-marketing}
\begin{tabular}{
    l                   
    l                   
    S[table-format=2.1] 
    S[table-format=2.1] 
    S[table-format=2.1] 
    S[table-format=2.1] 
}
\toprule
\textbf{Rank} & \textbf{Agent} & {\textbf{Average}} & {\textbf{E-commerce}} & {\textbf{APP}} & {\textbf{Game}} \\
\midrule
1  & o3                   & \textbf{50.8} & \textbf{50.6} & 46.0 & \textbf{52.5} \\
2  & Claude-3.7-Sonnet    & 47.6 & 46.2 & \textbf{46.9} & 43.5 \\
3  & Grok3-Search         & 46.5 & 45.3 & 52.2 & 47.5 \\
4  & Gemini-2.5-Pro       & 45.9 & 45.8 & 40.1 & 45.4 \\
5  & Gemini-2.5-Flash     & 45.3 & 40.8 & 45.3 & 46.7 \\
6  & o4-mini-high         & 43.5 & 44.1 & 39.0 & 40.5 \\
7  & Perplexity-Research  & 40.2 & 41.3 & 32.0 & 44.0 \\
8  & Perplexity-Search    & 34.4 & 36.6 & 26.4 & 31.1 \\
9  & GPT-4o               & 32.0 & 30.4 & 36.4 & 26.8 \\
\bottomrule
\end{tabular}
\label{tab:recruitment_benchmark}
\end{table}

\begin{table}[htbp]
\centering
\caption{Evaluation results on recruitment}
\label{tab:recruitment}
\begin{tabular}{
    l                   
    l                   
    S[table-format=2.1] 
    S[table-format=2.1] 
    S[table-format=2.1] 
    S[table-format=2.1] 
}
\toprule
\textbf{Rank} & \textbf{Agent} & {\textbf{Average}} & {\textbf{Company Map.}} & {\textbf{People-to-Info}} & {\textbf{Info-to-People}} \\
\midrule
1  & o3                   & \textbf{78.5} & \textbf{92.3} & \textbf{82.8} & \textbf{66.2} \\
2  & Perplexity-Search    & 64.4 & 88.5 & 59.4 & 51.9 \\
3  & Claude-3.7-Sonnet    & 61.4 & 86.5 & 47.2 & 54.8 \\
4  & o4-mini-high         & 61.4 & 71.3 & 50.1 & 62.9 \\
5  & Gemini-2.5-Flash     & 60.6 & 84.5 & 50.0 & 52.4 \\
6  & Perplexity-Research  & 59.1 & 88.5 & 56.3 & 41.4 \\
7  & Gemini-2.5-Pro       & 57.3 & 75.0 & 53.1 & 48.6 \\
8  & DeepSeek R1          & 48.3 & 75.0 & 42.2 & 34.8 \\
9  & Grok3-Search         & 47.1 & 82.8 & 45.0 & 13.3 \\
10 & GPT-4o               & 38.9 & 62.5 & 30.1 & 29.5 \\
\bottomrule
\end{tabular}
\label{tab:marketing_benchmark}
\end{table}

\section{Discussion}
\subsection{Evaluating Lifecycling Products with xbench-Index}
\label{sec:xbench_index}
Static benchmarks are susceptible to the test case leakage problem \citep{xu2024benchmarking}. The emergence of dynamic benchmarks like LiveBench \citep{white2024livebench} and LiveCodeBench \citep{jain2024livecodebench}, which augment evaluation sets with dynamically updated test cases, has mitigated this issue. However, new challenges persist in the evaluation of AI agent applications.

Firstly, \textbf{product versions of agent applications have lifecycles.} Agent products iterate rapidly, continuously integrating and developing new features, while older versions may be deprecated or taken offline. Although we can assess the capabilities of different products from similar agents at a given time, comparing the performance improvements of these products at different time points is challenging.

Secondly, \textbf{the external environments with which agents interact are also dynamic.} Even for the identical task, if its solution requires the use of tools with rapidly changing content, such as web applications, the test results will vary when conducted at different times.

\begin{table}[htbp]
\caption{Available results from the live evaluations of agents. Green cells stand for the available evaluation results, grey cells stand for unpublished versions, red cells stand for the historical offline agent versions.}
\resizebox{\linewidth}{!}{
    \begin{tabular}{@{}cllllllll@{}}
    \toprule
    Eval \textbackslash Versions & \multicolumn{1}{c}{APP1 v1} & \multicolumn{1}{c}{APP2 v1} & \multicolumn{1}{c}{APP3 v1} & \multicolumn{1}{c}{APP1 v2} & \multicolumn{1}{c}{APP2 v2} & \multicolumn{1}{c}{APP3 v2} & \multicolumn{1}{c}{APP1 v3} & \multicolumn{1}{c}{APP2 v3} \\ \midrule
    Eval 25-06                   & \cellcolor{myTableGreen}    & \cellcolor{myTableGreen}    & \cellcolor{myTableGreen}    & \cellcolor{myTableGrey}     & \cellcolor{myTableGrey}     & \cellcolor{myTableGrey}     & \cellcolor{myTableGrey}     & \cellcolor{myTableGrey}     \\
    Eval 25-08                   & \cellcolor{myTableRed}      & \cellcolor{myTableGreen}    & \cellcolor{myTableGreen}    & \cellcolor{myTableGreen}    & \cellcolor{myTableGreen}    & \cellcolor{myTableGrey}     & \cellcolor{myTableGrey}     & \cellcolor{myTableGrey}     \\
    Eval 25-10                   & \cellcolor{myTableRed}      & \cellcolor{myTableRed}      & \cellcolor{myTableGreen}    & \cellcolor{myTableGreen}    & \cellcolor{myTableGreen}    & \cellcolor{myTableGreen}    & \cellcolor{myTableGrey}     & \cellcolor{myTableGrey}     \\
    Eval 25-12                   & \cellcolor{myTableRed}      & \cellcolor{myTableRed}      & \cellcolor{myTableRed}      & \cellcolor{myTableRed}      & \cellcolor{myTableGreen}    & \cellcolor{myTableGreen}    & \cellcolor{myTableGreen}    & \cellcolor{myTableGreen}    \\ \bottomrule
    \end{tabular}
}
\label{tab:live_eval}
\end{table}

\Cref{tab:live_eval} illustrates the results available from live evaluations of agents. Although these results can be used to rank different concurrent products, the capability growth across different evaluation periods is not captured due to adjustments in the evaluation environment and tasks. Therefore, we aim to address the following question:

\textit{How can we design metrics to track the continuous growth of agent capabilities amidst the ongoing iteration of both evaluation sets and the agents themselves?}

Statistically, we can estimate the principal components of capabilities for each agent version from an incomplete score matrix. We employ Item Response Theory (IRT) \citep{hambleton1991fundamentals} to accomplish this estimation of agent capabilities. IRT models the ability of a test subject $\theta$, the item difficulty $b$, and the item discrimination index $a$ to predict the subject's score on a test item using the following model:

$$p(\theta) = \frac{1}{1 + e^{-a(\theta - b)}}$$

This formula ensures that the probability of a correct response $p$ ranges in $[0,1]$. A higher difficulty parameter $b$ decreases the probability of a correct response, whereas a higher ability parameter $\theta$ increases it. Items with a higher discrimination index $a$ typically exhibit a gentler slope in relation to ability $\theta$, indicating that the item can differentiate a wider range of subject capabilities.


\begin{figure}[htbp] 
  \centering 

  \newlength{\firstrowimageheight}
  \setlength{\firstrowimageheight}{5cm} 

  \begin{minipage}{\textwidth}
    \centering 
    
    \begin{subfigure}[t]{0.49\textwidth} 
      \centering 
      \includegraphics[height=\firstrowimageheight, width=\linewidth, keepaspectratio]{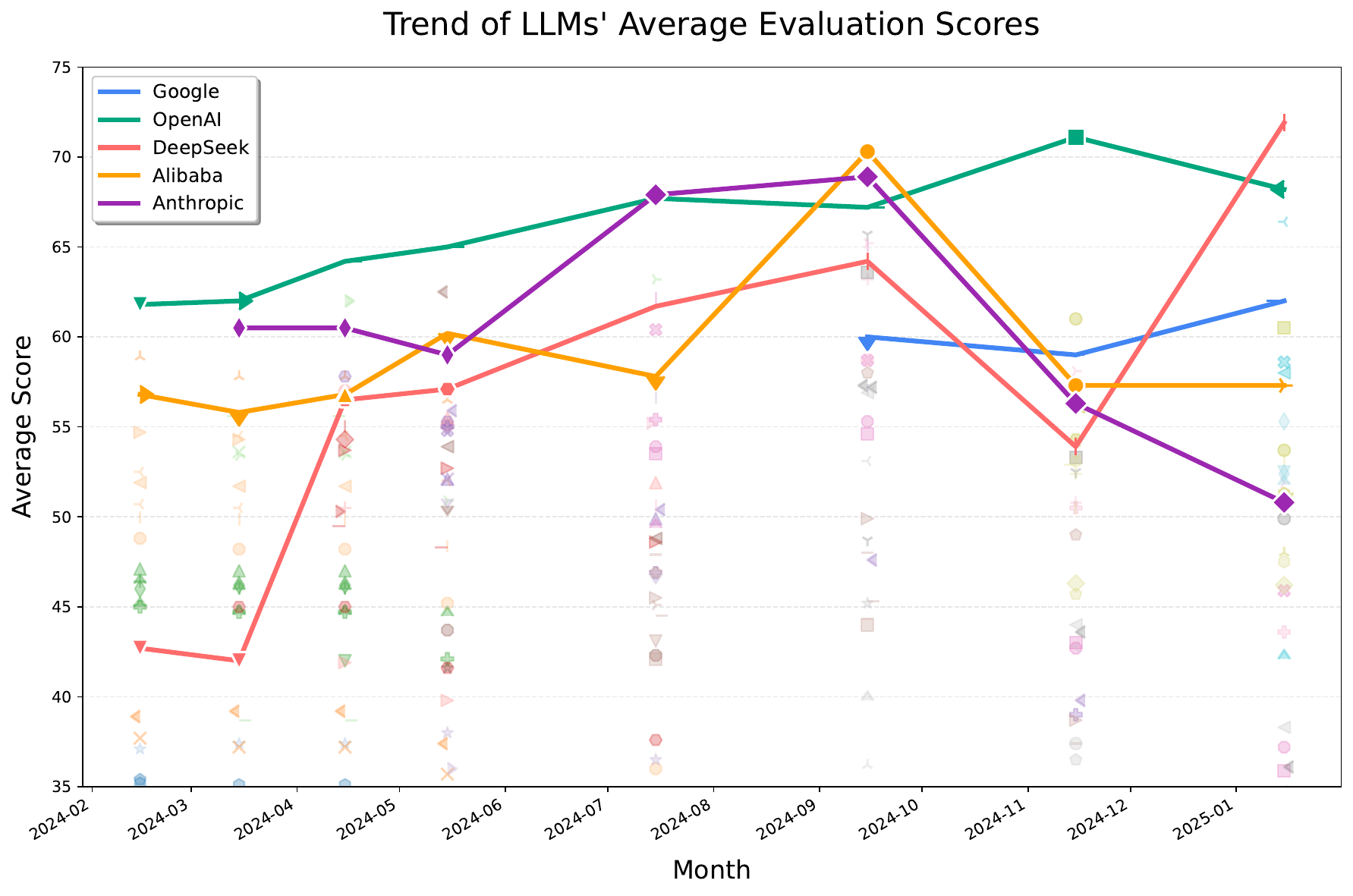} 
    \end{subfigure}
    \hfill 
    \begin{subfigure}[t]{0.49\textwidth} 
      \centering
      \includegraphics[height=\firstrowimageheight, width=\linewidth, keepaspectratio]{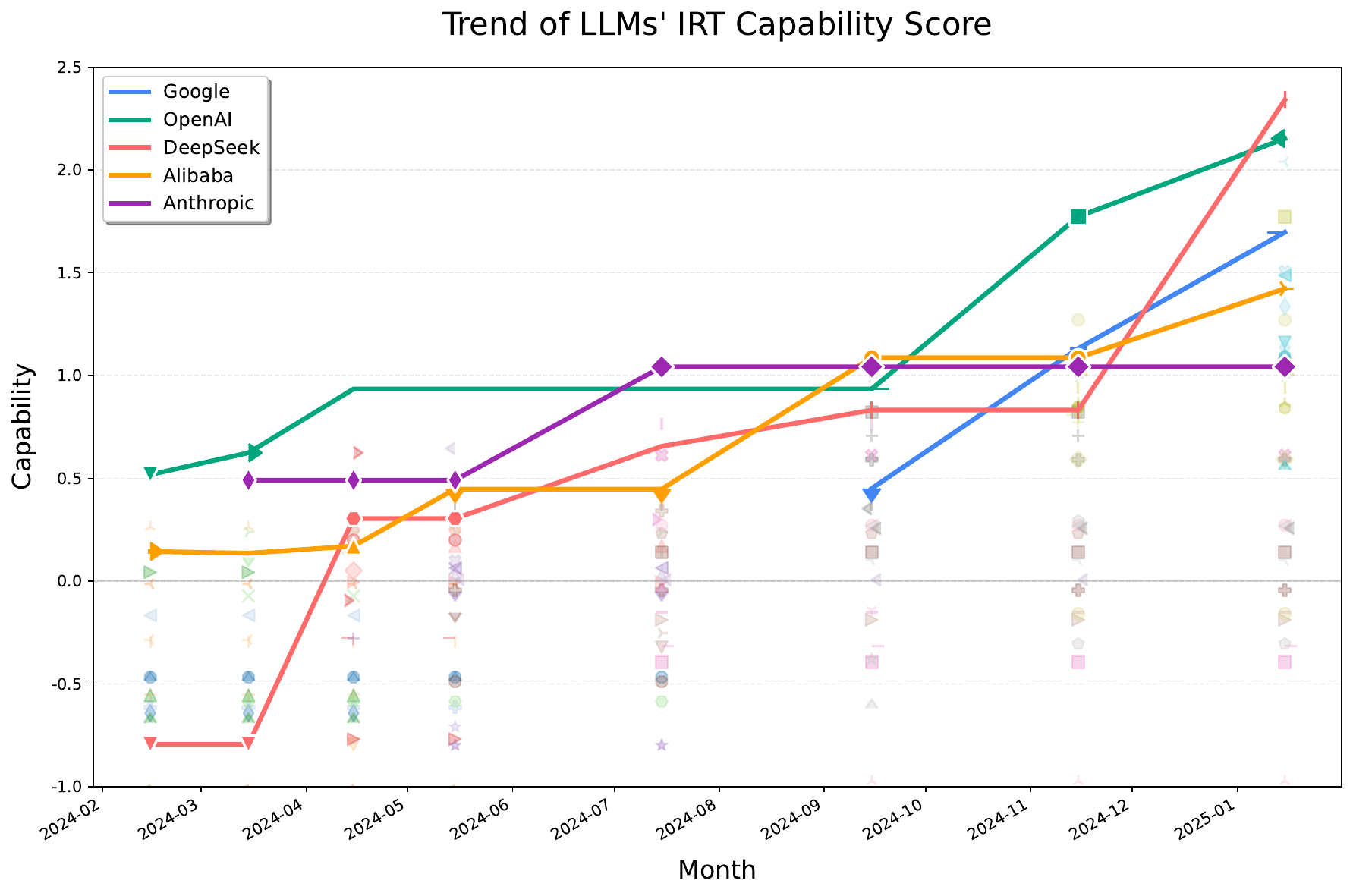} 
    \end{subfigure}
  \end{minipage}


  \begin{subfigure}{\textwidth} 
    \centering
    \includegraphics[width=\linewidth, keepaspectratio]{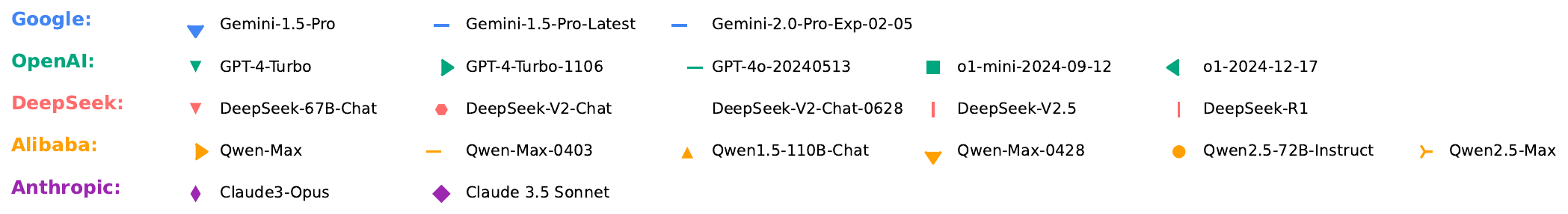} 
  \end{subfigure}
  \caption{OpenCompass origin evaluation and IRT capability estimation.} 
  \label{fig:opencompass_eval} 
\end{figure}

We utilize the dynamically updated OpenCompass \citep{2023opencompass} evaluation results \footnote{https://rank.opencompass.org.cn/leaderboard-llm/?m=25-01} to validate the IRT method. This leaderboard, initiated in February 2024, updates its question bank and releases evaluation results every 1-3 months. \Cref{fig:opencompass_eval} displays the scores of different models at various evaluation times; models from the same series are connected by lines of the same color. Although the leaderboard results effectively show the ranking of model capabilities at each evaluation point, the scores of models at different times are not directly comparable due to updates in the test items.

However, the capability scores estimated using IRT can effectively reflect the trend of continuous growth in model capabilities. We can observe the rapid advancement of Google's Gemini \citep{gemini2.5} model capabilities after October 2024, as well as two significant improvements brought about by the releases of DeepSeek-V2 \citep{liu2024deepseek} and its subsequent revision DeepSeek-r1 \citep{r1}.

In future agent evaluations, we will continuously report the IRT capability scores for different products on the agent evaluation set. This will enable the over time observation of signals related to \textbf{developmental velocity} and \textbf{key breakthroughs} that go beyond simple rankings.

\subsection{Evaluating Agents Tech-Market-Fit}

Cost is also one of the decisive factors in the practical adoption of agent applications.

Inference scaling enables models and agents to achieve superior performance through the investment of greater inference compute resources. This investment can manifest as longer chains of thought derived from reinforcement learning, or as further performance enhancements by introducing additional inference and summarization steps building upon these chains of thought.

However, in real-world applications, it is crucial to consider the cost-benefit ratio of inference scaling, striving for an optimal balance among cost, latency, and performance. Similar to ARC-AGI \citep{arcagi}, we aim to report for each evaluation set its demand curve, human capability curve, and the optimal supply curve of existing products on a performance-cost graph.

\begin{figure}[htbp] 
    \centering 

    \begin{subfigure}[b]{0.33\textwidth}
        \centering
        \includegraphics[width=\linewidth]{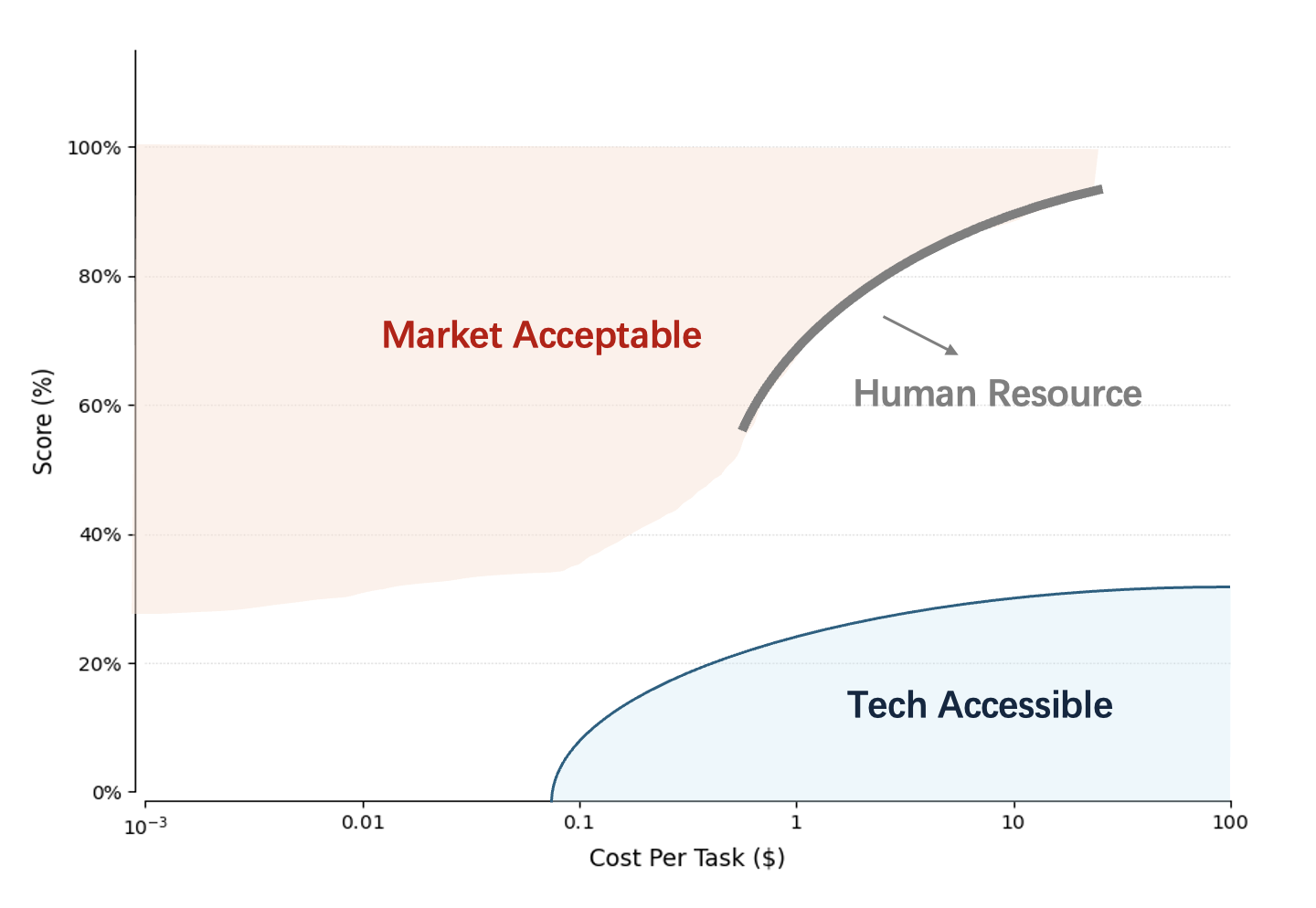}
        \caption{Stage 1. TMF Not Achieved}
        \label{fig:tmf_sub1} 
    \end{subfigure}%
    \hfill 
    \begin{subfigure}[b]{0.33\textwidth}
        \centering
        \includegraphics[width=\linewidth]{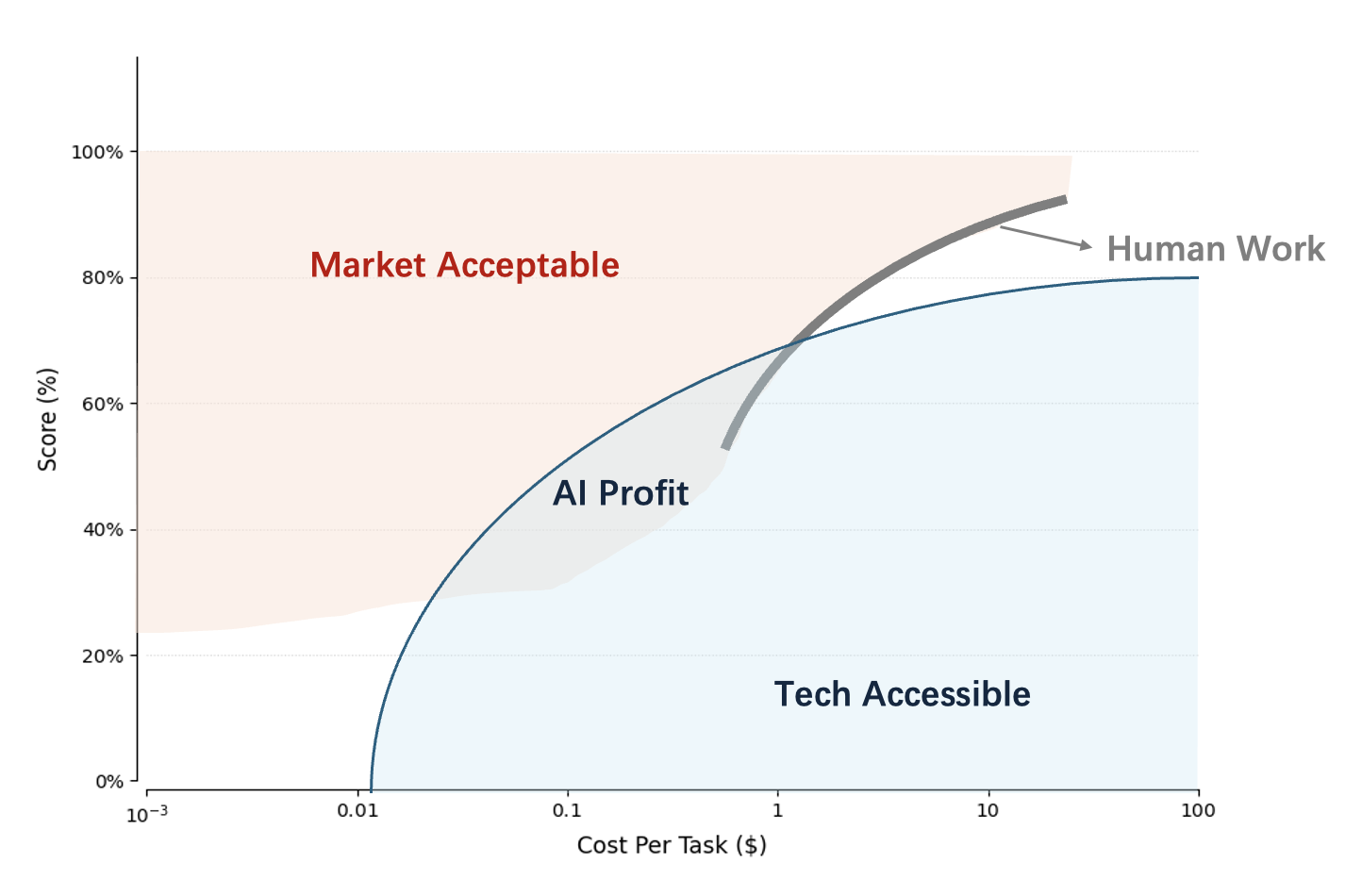}
        \caption{Stage 2. Collaboration}
        \label{fig:tmf_sub2} 
    \end{subfigure}%
    \hfill 
    \begin{subfigure}[b]{0.33\textwidth}
        \centering
        \includegraphics[width=\linewidth]{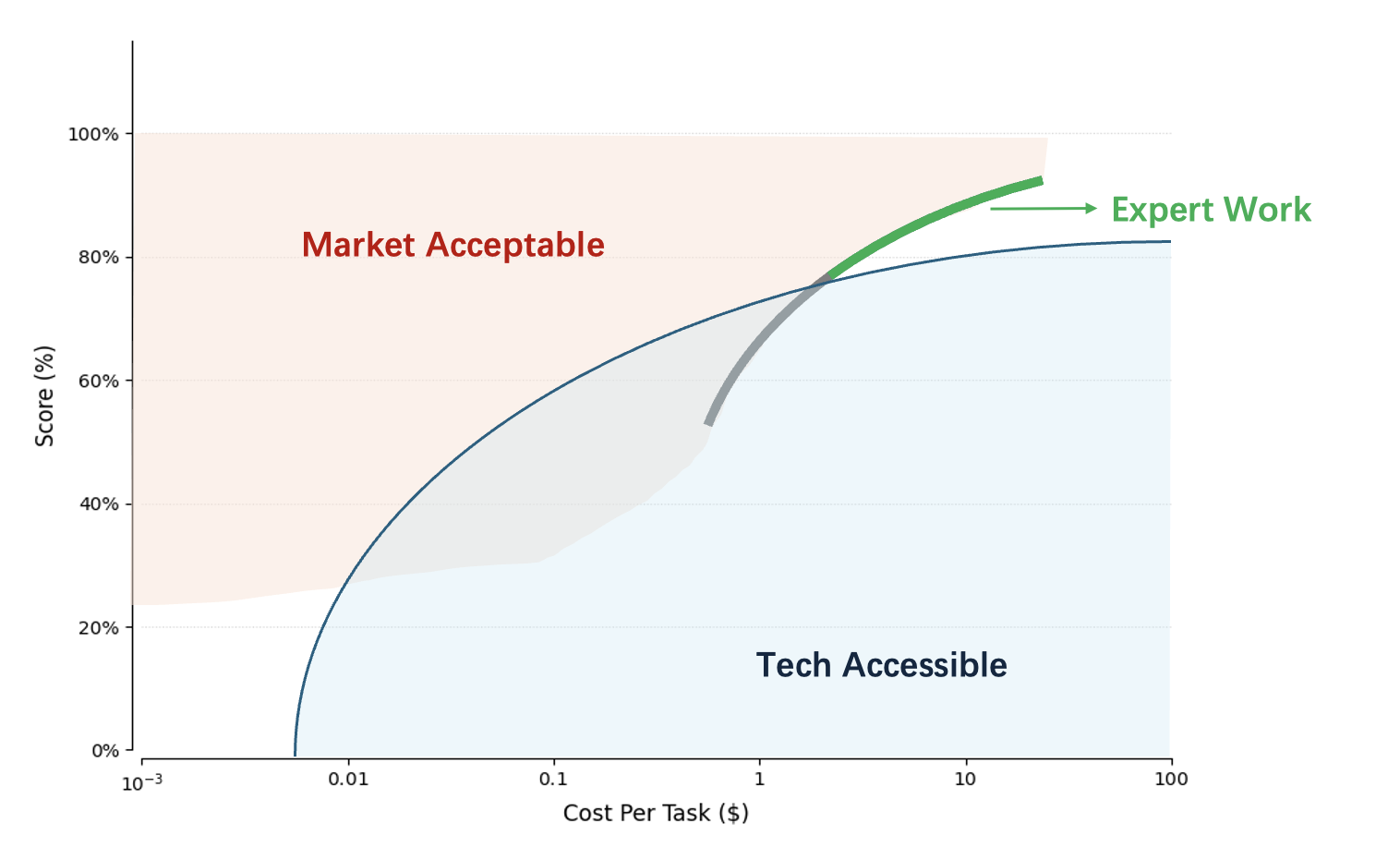}
        \caption{Stage 3. Specialized Agents}
        \label{fig:tmf_sub3} 
    \end{subfigure}

    \caption{Tree stages of Tech-Market-Fit.}
    \label{fig:tmf} 
\end{figure}

As shown in \cref{fig:tmf}, on the performance-cost graph, we can distinguish an upper-left region as the ``market acceptable region" and a lower-right region as the "tech accessible region". Human labor cost should constitute a part of the boundary of the market acceptable region. \Cref{fig:tmf_sub1} typically illustrates a state where the technology has not yet achieved practical adoption, whereas \cref{fig:tmf_sub2} depicts the state after achieving Tech-Market-Fit (TMF). The intersection area therein represents the incremental value brought by AI.

For AI applications that have achieved TMF, human resources should be increasingly allocated to the frontiers of the domain and to tasks that remain challenging for effective evaluation. Consequently, the market will re-price the value of human contributions, factoring in the differing scarcities of human expertise versus AI computational power (\Cref{fig:tmf_sub3}).

We believe that each professional domain will undergo three stages:

\textbf{Stage 1. Technology-Market Fit Not Achieved:} In this stage, the regions of tech accessible and market acceptable do not intersect. Agent applications are merely tools or concepts, incapable of delivering tangible results or generating value at scale. The impact of agents on human work is minimal.

\textbf{Stage 2. Agents and Humans Collaborating:} The tech accessible and market acceptable regions begin to overlap. This intersecting area represents the incremental value brought by AI, which includes: 1. Providing viable services at a cost lower than the minimum human labor cost, and 2. Assisting in handling repetitive tasks and works with moderate quality requirements. However, high-level work, characterized by data scarcity and greater difficulty, still necessitates human execution. At this juncture, due to the scarcity of such high-level human expertise, the AI profits obtained by enterprises may be allocated to compensate for this high-end human work output.

\textbf{Stage 3. Specialized Agents:} Domain experts take the lead in constructing evaluation systems and guiding the iteration of agents. The role of experts transitions from directly delivering results to building professional evaluation and training frameworks for specialized agents, and subsequently offering these specialized AI services at scale.

The transition from stage 1 to stage 2 is primarily driven by breakthroughs in AI technology, coupled with the scaling of compute and data. The progression from stage 2 to stage 3 is contingent upon the involvement of experts who are familiar with the specific requirements, standards, and have experience to the specific domains.

Furthermore, in certain domains, AI may introduce novel approaches to meet existing needs, thereby transforming established business processes and the composition of production relations.

While AI might lead to value shifts and alter the structure of human resource demands, we believe that society will ultimately experience an increase in overall human welfare due to the enhanced production efficiency and innovative business models fostered by AI.


\section{Related Works}
This work involves constructing real-world evaluations, utilizing an LLM Judge to score open-ended tasks, and aim to track and predict the growth of domain agents' capabilities. In this section, we briefly review existing works and their relevance.

\subsection{Real-world AI Evaluations}
Real-world evaluation aims to transform practical tasks and their execution environments into assessable benchmarks. Existing work has built evaluations for core AI capabilities, such as browser-use agents~\citep{deng2023mind2web}, GUI agents~\citep{xie2024osworld}, as well as for vertical domains like coding~\citep{jimenez2023swe}, customer service~\citep{yao2024taubench}, database operations~\citep{lei2024spider}, and healthcare~\citep{arora2025healthbench}. High-quality vertical benchmarks have driven the development of domain-specific agents and accelerated the emergence of commercially successful companies such as Cursor, Mercer, and Sierra. These existing vertical evaluations also pursue a profession-aligned model. We hope to broaden this profession-aligned evaluation paradigm, enhance the quality of evaluation tasks by collaborating with top industry experts, and commit to long-term updates and reporting of our results.

\subsection{Constitution AI for Judge and Reward Models}
As more real-world tasks require open-ended outputs, rule-based judgments have shown limitations in their applicability across different domains. Consequently, LLM-Judges based on scoring rubrics have been widely adopted~\citep{starace2025paperbench,lin2024wildbench,sirdeshmukh2025multichallenge,scaleai2025vista,fast2024autonomous}. Constitutional AI first applied rubric-based alignment for safety purposes, using human values to construct safety principles and alignment data~\citep{bai_constitutional_2022}; this approach was later extended to the training of general-purpose reward models~\citep{liu2025inference}. An increasing number of evaluation benchmarks have also extended the use of rubric-based LLMs to other domains. A profession-aligned evaluation requires the construction of a professional assessment framework. Some tasks in xbench adopt a rubric-based evaluation method where the scoring framework is co-developed with industry experts. The evaluation criteria can be tailored to individual tasks, categories of tasks, or entire workflows, depending on practical needs.

\subsection{Scaling Laws for LLMs' Performance}
Scaling laws have been a core principle driving research in AI capabilities~\citep{kaplan2020scalinglaws}, bringing predictability to the improvement of AI and effectively guiding the training process of large models. The scaling law for training loss revealed a power-law relationship between the final loss of an LLM and the compute invested, which in turn guided the selection of optimal parameter counts and data scale for training~\citep{hoffmann2022computeoptimal}. Scaling laws for hyperparameters have also been established based on experiments with smaller models~\cite{kadra2023scalinglawshyperparameteroptimization}. Further research has utilized scaling laws to predict performance on validation sets, test sets, and downstream tasks for better estimation of large model capabilities~\citep{isik2025scalinglawsdownstreamtask,xu2025unveilingdownstream}.

However, research on scaling laws that guide the evolution of agents remains relatively scarce. As mentioned in the literature, \cite{measuring-ai-ability-to-complete-long-tasks} observe that the effective time for AI to complete a task doubles every seven months. Past growth is primarily driven by improvements in foundational model capabilities, whereas future advancements are likely to come more from the design of agent work pipelines and from agents trained end-to-end in real-world environments. This paper focuses on the evaluation of vertical agents and the tracking of their productivity value scaling. We will continuously collect evaluation results for vertical agents to observe and track the characteristics of productivity scaling, and we will publish these findings as part of our future work.

\section{Summary}
In this work, we introduce xbench, a new evaluation suite designed to measure the real-world productivity of AI agents by aligning benchmarks with professional workflows. Moving beyond abstract technical skills, xbench uses live, expert-defined tasks from commercially significant fields to assess an agent's ability to deliver tangible business value.

As our initial implementations, we developed benchmarks for the Recruitment domain, evaluating agents' effectiveness in headhunting tasks like talent sourcing, and for Marketing, where we assess the ability to identify suitable influencers for real-world campaigns. We have established baseline results for leading contemporary agents on these tasks. To handle the dynamic nature of both agents and their environments, xbench is designed as a continuously updated system that uses IRT to track true capability growth over time. Our goal is to provide a clear, value-oriented framework for guiding and predicting the development of effective, domain-specific AI agents. 

\section{Acknowledgement}
We are deeply grateful to Yangjun Ruan for providing guidance on the design and writing of this article.
\bibliography{iclr2025_conference}
\bibliographystyle{iclr2025_conference}

\newpage
\appendix
\section{Appendix}
\subsection{Prompts for response collection and evaluation}
\label{appdix:recruit_prompts}
\begin{table*}[htbp]
\centering
\begin{minipage}{0.99\linewidth}\vspace{0mm}    \centering

\begin{tcolorbox}[colframe=black!75!white, colback=white, coltitle=white, title=Prompt for Recruitment Agents' Response Collection, fonttitle=\bfseries]
\small
\textbf{Prompt for Jd Analysis:} \vspace{0.1cm} \\
You are a recruitment expert. You need to accurately identify target \texttt{\{search object\}} based on the job requirements description.
\vspace{0.1cm} \\
Job requirements: \texttt{\{question\}}
\vspace{0.1cm} \\
You need to pay attention to:
\begin{enumerate}[leftmargin=*]
    \item Unless otherwise specified, only consider candidates within China.
    \item Do not over-search. If the number of returned results exceeds the actual demand, we will truncate to the specified number of objects at the beginning for evaluation.
\end{enumerate}
At the end of your analysis, you need to return in the following format:
\vspace{0.1cm} \\
\texttt{\#\# Search Results}

\texttt{Search Object 1: xx, xx}

\texttt{Search Object 2: xx, xx, xx}

\noindent\makebox[\linewidth]{\tikz[baseline]{\draw[dashed] (0,0) -- (\linewidth,0);}}\vspace{2mm}
\textbf{Prompt for People-to-Info:} 
\vspace{0.1cm} \\
You are a talent information search specialist. Based on the following background and reference information, please search for a comprehensive yet concise background and experience of this individual:
\vspace{0.1cm} \\
Background: \texttt{\{Background\}}
\vspace{0.1cm} \\
Reference Information: \texttt{\{Reference Information\}}
\vspace{0.1cm} \\
You need to note:
\begin{enumerate}[leftmargin=*]
    \item Unless otherwise specified, only consider the background information of Chinese individuals.
    \item The target individual is unique; the reference information is to help you pinpoint the target individual.
    \item We will prepare verification questions regarding the target individual, and the LLM will use the information you provide to attempt to answer them.
\end{enumerate}

\noindent\makebox[\linewidth]{\tikz[baseline]{\draw[dashed] (0,0) -- (\linewidth,0);}}\vspace{2mm}
\textbf{Prompt for Info-to-People:} \vspace{0.1cm} 
\\
You are a talent information search specialist. Based on the following reference information about talent, please identify the target person (or people, as needed):
\vspace{0.1cm} \\
Question: \texttt{\{question\}}
\vspace{0.1cm} \\
You need to note: 
\begin{enumerate}[leftmargin=*]
    \item Unless otherwise specified, only consider individuals from \texttt{\{country\}}, or \texttt{\{type of person\}}.
\end{enumerate}
\end{tcolorbox}
\caption{Prompt for recruitment agents' response collection.}

\label{tab:collection v1.0}
\end{minipage}
\end{table*}

\begin{table*}[t]
\centering
\begin{minipage}{0.99\linewidth}\vspace{0mm}    \centering

\begin{tcolorbox}[colframe=black!75!white, colback=white, coltitle=white, title=Prompt for Recruitment Response Evaluation, fonttitle=\bfseries]
\small
\textbf{Evaluation Prompt for Jd Analysis:} \vspace{0.1cm} \\
Please act as a recruitment evaluation expert. Strictly follow the judgment criteria below to score and briefly comment on the model's response:
\vspace{0.1cm} \\
Problem description: \texttt{\{question\}}
\vspace{0.1cm} \\
Standard answer: \texttt{\{ground truth answer\}}
\\\\
Candidate's response: \texttt{\{answer\}}
\vspace{0.1cm} \\
Briefly explain the reasons and provide a score.
\\\\
Evaluation Method:
\begin{enumerate}[leftmargin=*]
    \item First, extract \texttt{\{search object\}} from the results, and summarize the results for each type of \texttt{\{search object\}}.
    \item Determine the count for each type of \texttt{\{search object\}} in the extracted answers.
    \item For each type of \texttt{\{search object\}}, truncate the list of results if it is longer than the quantity in the standard answer.
    \item For each piece of information in the standard answer, check if it is covered by the information provided by the AI. If covered, mark as True; otherwise, mark as False. Calculate the coverage of the standard answer.
    \item For content that is too long or mismatched, check if it is fabricated and analyze its hallucinatory nature. Your final score should comprehensively consider coverage, hallucination, and information quality, etc.
\end{enumerate}

Then you need to consider scoring the results. Your scoring should be as strict and differentiating as possible.
\\\\
Scoring Rubric:
\vspace{0.2cm} \\
1 point: The provided industry, company, school, or department information does not match the standard answer; the provided information is basically worthless.
\vspace{0.2cm} \\
2 points: The provided information covers 50\% of the reference answer content, but the judgment criteria are not met, and there are unacceptable hallucinations.
\vspace{0.2cm} \\
3 points: The provided information covers 85\% of the reference answer content, judgment criteria are partially met, and a small amount of information is untrue or unreasonable.
\vspace{0.2cm} \\
4 points: The provided information covers 95\% of the reference answer content, judgment criteria are basically met, and there are almost no unacceptable hallucinations.
\vspace{0.2cm} \\
5 points: The provided information covers 100\% of the reference answer content, the judgment criteria are met, and there are no unacceptable hallucinations.
\\\\
You must provide the score in the last line. The scoring range is 1-5 points, with 1 being the lowest and 5 the highest. The output format for the last line is:
\vspace{0.1cm} \\
\texttt{"Score: X"}

\end{tcolorbox}
\caption{Prompt for recruitment response evaluation (Part 1).}

\label{tab:collection v1.2}
\end{minipage}
\end{table*}

\begin{table*}[t]
\centering
\begin{minipage}{0.99\linewidth}\vspace{0mm}    \centering

\begin{tcolorbox}[colframe=black!75!white, colback=white, coltitle=white, title=Prompt for Recruitment Response Evaluation, fonttitle=\bfseries]
\small
\textbf{Evaluation Prompt for People-to-Info:} \vspace{0.1cm} \\
Please act as a talent information evaluation expert. Based on the scoring criteria below, score and briefly comment on the candidate's response:
\\\\
Background: {background}
\vspace{0.1cm} \\
Reference Information: \texttt{\{reference information\}}
\vspace{0.1cm} \\
Verification Questions: \texttt{\{verification questions\}}
\vspace{0.1cm} \\
Standard Answers: \texttt{\{ground truth answers\}}
\vspace{0.1cm} \\
Scoring Criteria: \texttt{\{scoring criteria\}}
\vspace{0.1cm} \\
Based on AI-provided information: \texttt{\{AI searched results\}}
\\\\
Please provide a score and briefly explain the reasons.

You need to analyze each verification question and score the information provided by the AI by comparing it with the standard answers. The main focus is on whether the AI-provided information perfectly answers the verification questions. The scoring range is 1-5 points, with 1 being the lowest and 5 the highest.
\\\\
Scoring Rubric:
\vspace{0.2cm} \\
1 point: The industry, company, school, or department information provided by the AI does not match the standard answers, or it consists mostly of fabricated, non-existent experiences.
\vspace{0.2cm} \\
2 points: The information provided by the AI covers 20\% of the reference answer content, but the judgment criteria are not met, and there are unacceptable hallucinations.
\vspace{0.2cm} \\
3 points: The information provided by the AI covers 50\% of the reference answer content, judgment criteria are partially met, and a small amount of information is untrue or unreasonable.
\vspace{0.2cm} \\
4 points: Most of the verification questions can be correctly answered by the AI-provided information, and there are no hallucinated experiences.
\vspace{0.2cm} \\
5 points: All verification questions can be answered by the AI-provided information, consistent with the standard answers, and no hallucination issues occur.
You must provide the final score in the last line. The scoring range is 1-5 points, with 1 being the lowest and 5 the highest.
\\\\
The output format for the last line is:
\vspace{0.1cm} \\
\texttt{"Score: X"}

\end{tcolorbox}
\caption{Prompt for recruitment response evaluation (Part 2).}

\label{tab:collection v1.3}
\end{minipage}
\end{table*}

\begin{table*}[t]
\centering
\begin{minipage}{0.99\linewidth}\vspace{0mm}    \centering

\begin{tcolorbox}[colframe=black!75!white, colback=white, coltitle=white, title=Prompt for Recruitment Response Evaluation, fonttitle=\bfseries]
\small
\textbf{Evaluation Prompt for Info-to-People:} \vspace{0.1cm} 
\\
Please act as a corporate information evaluation expert. Based on the verification method below, score and briefly comment on the candidate's response:
\\\\
Question: \texttt{\{question\}}
\vspace{0.1cm} \\
Standard Answer: \texttt{\{ground truth answer\}}
\vspace{0.1cm} \\
Verification Method: \texttt{\{verification method\}}
\\\\
Candidate's Response: \texttt{\{answer\}}
\\\\
Please provide a score from 1-5 and briefly explain the reasons.
\vspace{0.1cm} \\
You must provide the final score in the last line. The scoring range is 1-5 points, with 1 being the lowest and 5 the highest.
\\\\
Scoring Rubric:
\vspace{0.2cm} \\
1 point: The answer is completely incorrect, with a lot of hallucinated information.
\vspace{0.2cm} \\
2 points: The answer is incorrect, but there is less hallucinated information, or it admits to not finding the information.
\vspace{0.2cm} \\
3 points: The answer is partially correct, with less hallucinated information.
\vspace{0.2cm} \\
4 points: The answer is mostly correct, possibly with a few errors or some hallucinated information.
\vspace{0.2cm} \\
5 points: The final selection is consistent with the standard answer, and no other task information is hallucinated.
\\\\
The output format for the last line is:
\vspace{0.1cm} \\
\texttt{"Score: X"}

\end{tcolorbox}
\caption{Prompt for recruitment response evaluation (Part 3).}

\label{tab:collection v1.4}
\end{minipage}
\end{table*}

\begin{table*}[t]
\centering
\begin{minipage}{0.99\linewidth}\vspace{0mm}    \centering

\begin{tcolorbox}[colframe=black!75!white, colback=white, coltitle=white, title=Prompt for Marketing Agents’ Response Collection, fonttitle=\bfseries]
\small
\textbf{Prompt for Marketing:} \vspace{0.1cm} 
\\
You are a professional market operator. You now need to find bloggers who meet the requirements for advertising based on client needs.
\vspace{0.1cm} \\
The range of platforms you can access is \texttt{\{platfrom\_list\}}.
\vspace{0.1cm} \\
The client's category is \texttt{\{type\}}, the product name is \texttt{\{product\}}, and you need to find a total of \texttt{\{topk\}} bloggers. The client's requirements form is as follows:
\\\\
\texttt{\{demand\}}
\\\\
Now you need to return a JSON formatted list of bloggers, with the format shown below:
\begin{verbatim}
[
    {
        "Blogger Name": "Blogger Name",
        "Blogger Link": "Blogger Link",
    },
    {
        "Blogger Name": "Blogger Name",
        "Blogger Link": "Blogger Link",
    },
    ...
]
\end{verbatim}
% ...

Here, an example of a blogger link is "https://www.youtube.com/@TED", and the blogger name is "TED". You must provide the website link, as this is a key basis for evaluation. You should place the bloggers you are more optimistic about at the beginning.

\end{tcolorbox}
\caption{Prompt for marketing agents’ response collection.}

\label{tab:collection v1.5}
\end{minipage}
\end{table*}

\begin{table*}[t]
\centering
\begin{minipage}{0.99\linewidth}\vspace{0mm}    \centering

\begin{tcolorbox}[colframe=black!75!white, colback=white, coltitle=white, title=Prompt for Marketing Response Evaluation, fonttitle=\bfseries]
\small
\textbf{Prompt for Single Influencer Info Summary:} \vspace{0.1cm} 
\\
You are a market operations expert who can analyze a blogger's information summary and the client's reasons for selection based on user needs and the client's chosen blogger. Below you will see blogger information and a client demand form. You need to provide an analysis and summary of this information.
\\\\
Client demand information: 
\vspace{0.1cm} 
\texttt{\{consumer\_demands\}}
\\\\
Blogger information:
\vspace{0.1cm} 
\texttt{\{profile\}}

\noindent\makebox[\linewidth]{\tikz[baseline]{\draw[dashed] (0,0) -- (\linewidth,0);}}\vspace{2mm}
\textbf{Prompt for Full Influencer Info:} \vspace{0.1cm} 
\\
You are a market operations expert who can analyze a blogger's information summary and the client's reasons for selection based on user needs and the client's chosen blogger. Below, you will see a compilation of multiple bloggers' information and a client demand form. You need to construct an analytical persona of an ideal blogger; its detailed attributes should be repeated as much as possible.
\vspace{0.1cm} \\
Client demand information:
\vspace{0.1cm} \\
\texttt{\{consumer\_demands\}}
\\
{
\scriptsize
\begin{verbatim}
"Blogger Information:" + "\nBlogger Information:".join(summary_list))
\end{verbatim}
}

\noindent\makebox[\linewidth]{\tikz[baseline]{\draw[dashed] (0,0) -- (\linewidth,0);}}\vspace{2mm}
\textbf{Prompt for Consumer Demands Decompose:} \vspace{0.1cm} 
\\
Please analyze the following demand form, analyze the conditions for selecting advertising bloggers, classify them into necessary conditions and flexible conditions, and assign a weight (1-10 points) to each condition.
\vspace{0.1cm} \\
Note:
\begin{enumerate}[leftmargin=*]
    \item Information related to dates, times, etc., are not conditions; please ignore them.
    \item Conditions are selected based on blogger characteristics, not sales strategies or product information.
    \item Please ignore any blank fields that are not filled in.
    \item Each condition must be given a specific weight score.
    \item Please return the response in JSON format, as shown below:
\end{enumerate}
{
\begin{verbatim}
{
    "Necessary Conditions": [
    {
        "Condition": "Condition Description", 
        "Weight": Score, 
        "Reason": "Why this is a necessary condition"
    },
    ...
    ],
    "Flexible Conditions": [
    {
        "Condition": "Condition Description", 
        "Weight": Score, 
        "Reason": "Why this is a flexible condition"
    },
    ...
    ]
}
\end{verbatim}
}
Demand form content:
\vspace{0.1cm} \\
\texttt{\{requirements\_text\}}

\end{tcolorbox}
\caption{Prompt for marketing response evaluation (Part 1).}

\label{tab:collection imagev1.6}
\end{minipage}
\end{table*}

\begin{table*}[t]
\centering
\begin{minipage}{0.99\linewidth}\vspace{0mm}    \centering

\begin{tcolorbox}[colframe=black!75!white, colback=white, coltitle=white, title=Prompt for Marketing Response Evaluation, fonttitle=\bfseries]
\small
\textbf{Evaluation Prompt for Influencer Eval:} \vspace{0.1cm} 
\\
You are a market operations expert who can provide a suitability score for a potential candidate blogger based on user needs and an ideal blogger persona summarized from client-selected bloggers. You need to comprehensively consider each element of the ideal blogger persona, give a score of 1-5 points for each element, and finally provide an overall suitability score of 1-5 points, then determine whether to select this blogger. 
\vspace{0.1cm} \\
Client demand information:
\vspace{0.1cm} \\
\texttt{\{demand\}}
\\\\
Client demand analysis: \texttt{\{demand\_analysis\}}
\vspace{0.1cm} \\
Ideal blogger persona summary: \texttt{\{profile\_standard\}}
\vspace{0.1cm} \\
Link of the blogger to be evaluated: \texttt{\{influencer\_link\}}
\vspace{0.1cm} \\
Information of the blogger to be evaluated: 
\texttt{\{condidate\_profile\}}
\\\\
Output format:
\vspace{0.1cm} \\
You need to output in a JSON format, containing four parts: 1. Analysis, a string representing the comprehensive scoring rationale, 2. Detailed Scoring, a dictionary where each key is the name of a scoring point and the value is the score for that point (1-5), 3. Overall Score, 1-5 points, 4. Selected, a string which can only be 'Yes' or 'No'. 
\vspace{0.1cm} \\
For example:
\begin{verbatim}
{
    "Analysis": "xxxx",
    "Detailed Scoring": {
        "Scoring Point 1": x,
        "Scoring Point 2": x,
        "Scoring Point 3": x,
        "Scoring Point 4": x
    },
    "Overall Score": x,
    "Selected": "Yes/No"
}
\end{verbatim}

Please note that your scoring criteria must be highly differentiating. Bloggers selected by the client must meet all requirements as much as possible. The design of scoring points should not be very broad; they should be specific and differentiating. Your final scoring result should be able to clearly select bloggers, rather than providing ambiguous judgments among several 'pretty good' bloggers.

\end{tcolorbox}
\caption{Prompt for marketing response evaluation (Part 2).}

\label{tab:collection image v1.7}
\end{minipage}
\end{table*}

\subsection{Complete Examples of the Evaluation Tasks}
\begin{table}[htbp]
\caption{The complete example of company mapping task.}
\begin{small}
\resizebox{\linewidth}{!}{
    \begin{tabular}{@{}ll@{}}
    \toprule
    \textbf{Task Type} &
      Company Mapping \\ \midrule
    \textbf{\begin{tabular}[c]{@{}l@{}}Task\\ Description\end{tabular}} &
      \begin{tabular}[c]{@{}l@{}}Target Candidate Profile Analysis Based on Job Requirements\\      \\ \textbf{Job Analysis:}\\ This role is an Influencer Marketing Specialist for an ACG (Animation, Comics, \\ Games) AIGC (AI Generated Content) product. The core requirement is to conduct \\ KOL (Key Opinion Leader) / KOC (Key Opinion Consumer) marketing on mainstr-\\eam social media platforms to drive product growth.\\      \\ \textbf{Target Talent Profile:}\\      \\ \textbf{Core Background Requirements:}\\ - Bachelor's degree or higher, with a minimum of three years of experience in influ-\\encer marketing. \\      - Prior experience in a growth-focused role at a major internet platform is preferred. \\      - In-depth understanding of the commercialization strategies of mainstream short-\\video and social media platforms. \\       \\      \textbf{Professional Competencies:}\\- Demonstrated ability in astute influencer identification and discerning trending \\content.\\      - Proven track record of successful viral influencer marketing campaigns.\\      - Comprehensive knowledge of ACG culture and a nuanced understanding of the \\user psychology for        male-oriented and female-oriented content. \\      - Proficiency in creative planning, business development (BD) negotiation, and pro-\\ject management.\\      \\ \textbf{Personal Traits:}\\ - Highly extroverted (e.g., "Mega E" personality type) or exceptionally engaging \\individual. \\      - Extremely self-driven with an entrepreneurial mindset.\\      - Possesses a strong interest and aspiration in content dissemination.\\      - Open-minded and unconstrained by niche cultural limitations (unfettered by ``circle \\culture").\end{tabular} \\ \midrule
    \textbf{\begin{tabular}[c]{@{}l@{}}Verifier\\ Answers\end{tabular}} &
      \begin{tabular}[c]{@{}l@{}}Seasoned professionals J and K from marketing agencies, specialized MCN organiz-\\ations G, H, and I, in addition to the marketing and growth divisions E and F of social\\ media platforms, e-commerce platforms C and D, and short-video platforms A and B.\end{tabular} \\ \bottomrule
    \end{tabular}
}
\end{small}
\label{tab:case_full_company_mapping}
\end{table}

\begin{table}[ht]
\caption{The complete example of People-to-Info task.}
\resizebox{\linewidth}{!}{
    \begin{tabular}{@{}ll@{}}
    \toprule
    \textbf{Task Type} &
      People-to-Info \\ \midrule
    \textbf{\begin{tabular}[c]{@{}l@{}}Task\\ Discription\end{tabular}} &
      \begin{tabular}[c]{@{}l@{}}Seeking comprehensive information for Ming Li, Backend Developer at Company A.     \\ \textbf{Reference Information:}     \\ \textbf{Work Experience}     \\     - August 2023 – Present: Company A, Java Backend Developer\\     - January 2022 – July 2023: Company B, Golang Backend Developer\\     - March 2020 – December 2021: Company C, Java Backend Developer\\     - June 2018 – February 2020: Company D, Java Backend Developer\\ \textbf{Education Background}\\     - 2014 – 2018: University A, Bachelor of Science in Computer Science\end{tabular} \\ \midrule
    \textbf{\begin{tabular}[c]{@{}l@{}}Verifier\\ Questions\end{tabular}} &
      \begin{tabular}[c]{@{}l@{}}\textbf{Question 1:} What were his key technical achievements during his tenure at Company A?\\ \textbf{Question 2:} What were the core projects he developed at Company B?\\ \textbf{Question 3:} Summarize his technical expertise and industry experience.\end{tabular} \\ \midrule
    \textbf{\begin{tabular}[c]{@{}l@{}}Verifier\\ Answers\end{tabular}} &
      \begin{tabular}[c]{@{}l@{}}\textbf{Answer 1:}\\      1) Responsible for the backend optimization of the short-form video recommendation\\      system, which increased the click-through rate by 18\%.\\      2) Developed real-time interactive features for live streaming rooms, supporting tens \\      of millions of concurrent users.\\      3) Constructed a user profiling and analytics platform.\\ \textbf{Answer 2:}\\      1) Led the architectural redesign of the music playback service, reducing latency \\      by 40\%.\\      2) Developed a personalized playlist recommendation algorithm.\\      3) Optimized the high-concurrency processing capabilities of the comment system.\\ \textbf{Answer 3:}\\      Technical Expertise: Java/Golang development, distributed systems, microservices\\      architecture.\\      Industry Experience: Financial payments, short-form video, music entertainment, \\      and social media sectors.\end{tabular} \\ \midrule
    \textbf{Time Cost} &
      30 min \\ \bottomrule
    \end{tabular}
}
\label{tab:case_full_people_to_info}
\end{table}


\end{document}

%% file: math_commands.tex
\usepackage{amsmath,amsfonts,bm}

\def\eqref#1{equation~\ref{#1}}

\def\1{\bm{1}}

\DeclareMathAlphabet{\mathsfit}{\encodingdefault}{\sfdefault}{m}{sl}
\SetMathAlphabet{\mathsfit}{bold}{\encodingdefault}{\sfdefault}{bx}{n}

